\documentclass[runningheads]{llncs}
\usepackage{graphicx}
\usepackage{comment}
\usepackage{amsmath,amssymb}
\usepackage{color}
\usepackage{siunitx}
\usepackage{booktabs}
\usepackage{bm}
\usepackage{mathtools}
\usepackage{xcolor}
\usepackage{pgfplots}
\usepackage{floatrow}
\usepackage{multirow}
\usepackage{wrapfig,lipsum,booktabs}
\usepackage{subcaption}
\usepackage{sidecap}
\usepackage[font=footnotesize,labelfont=bf]{caption}
\usepackage{url}

\usepackage{xspace}
\usepackage[export]{adjustbox}

\usetikzlibrary{arrows.meta}

\usepackage{bold-extra}

\newcommand*{\eg}{\emph{e.g.}\@\xspace}
\newcommand*{\ie}{\emph{i.e.}\@\xspace}

\newcommand*{\etal}{\emph{et al.}\@\xspace}

\newcommand{\refsec}[1]{Sec.\,\ref{sec:#1}}

\newcommand{\refequ}[1]{Eq.\,\ref{eq:#1}}
\newcommand{\reffig}[1]{Fig.\,\ref{fig:#1}}
\newcommand{\reftab}[1]{Tab.\,\ref{tab:#1}}

\makeatletter
\newcommand*{\etc}{%
    \@ifnextchar{.}%
        {etc}%
        {etc.\@\xspace}%
}

\definecolor{ubpubColor}{rgb}{0.43, 0.5, 0.5}
\definecolor{backboneColor}{rgb}{0.423, 0.325, 0.365}
\definecolor{fpnColor}{rgb}{0.255, 0.498, 0.416}

\newcommand{\PAR}[1]{\vskip4pt \noindent {\bf #1~}}
\newcommand{\PARbegin}[1]{\noindent {\bf #1~}}

\newcommand{\UNPUB}[1]{\textcolor{ubpubColor}{#1}}

\newcolumntype{P}[1]{>{\centering\arraybackslash}p{#1}}

\newcommand{\matrixvar}[1]{\bm{\mathit{#1}}}
\begin{document}

\pagestyle{headings}
\mainmatter
\def\ECCVSubNumber{1299}

\newcommand\blfootnote[1]{%
  \begingroup
  \renewcommand\thefootnote{}\footnote{#1}%
  \addtocounter{footnote}{-1}%
  \endgroup
}

\title{STEm-Seg: Spatio-temporal Embeddings for Instance Segmentation in Videos}

\titlerunning{STEm-Seg}
\author{Ali Athar*\inst{1} \and
Sabarinath Mahadevan*\inst{1} \and
Aljo\u{s}a O\u{s}ep\inst{2}  \and
Laura Leal-Taix\'{e}\inst{2}  \and
Bastian Leibe\inst{1}
}
\authorrunning{Athar, Mahadevan~\etal}

\institute{RWTH Aachen University, Germany 
\email{\{athar,mahadevan,leibe\}@vision.rwth-aachen.de}\\ \and
Technical University of Munich, Germany\\
\email{\{aljosa.osep,leal.taixe\}@tum.de}}

\maketitle

\blfootnote{* Equal contribution}
\begin{abstract}

%----------------------------ADDING EMBEDDING REPRESENTATION NOVELTY--------------

% ------What existing methods do---------
Existing methods for instance segmentation in videos typically involve multi-stage pipelines that follow the tracking-by-detection paradigm and model a video clip as a sequence of images. 
% ------What problems existing methods face--------
Multiple networks are used to detect objects in individual frames, and then associate these detections over time.
% Multiple networks are used to detect objects in individual frames, and subsequently associate them over time.
Hence, these methods are often 
%slow, 
non-end-to-end trainable and highly tailored to specific tasks. 
% ---- What do we do instead ---------
In this paper, we propose a different approach that is well-suited to a variety of tasks involving instance segmentation in videos.
In particular, we model a video clip as a single 3D spatio-temporal volume, and propose a novel approach that segments and tracks instances across space and time in a single stage. 
Our problem formulation is centered around the idea of spatio-temporal embeddings which are trained to cluster pixels belonging to a specific object instance over an entire video clip.
To this end, we introduce (i) novel mixing functions that enhance the feature representation of spatio-temporal embeddings, and (ii) a single-stage, proposal-free network that can reason about temporal context. Our network is trained end-to-end to learn spatio-temporal embeddings as well as parameters required to cluster these embeddings, thus simplifying inference.
% mitigating the need to use slow, often parametric clustering algorithms.
%
Our method achieves state-of-the-art results across multiple datasets and tasks. Code and models are available at \url{https://github.com/sabarim/STEm-Seg}. %We will make our code publicly avalible upon acceptance.

\end{abstract}

\section{Introduction}

The task of segmenting and tracking multiple objects in videos is becoming increasingly popular due to a surge in development of autonomous vehicles and robots that are able to perceive and accurately track surrounding objects.  
These advances are driven by the recent emergence of new datasets~\cite{Caelles19arXiv,Voigtlaender19CVPR,Yang19ICCV} containing videos with dense, per-pixel annotations of object instances.
The underlying task tackled in these datasets can be summarized as follows: given an input video containing multiple objects, each pixel has to be uniquely assigned to a specific object instance or to the background.

\begin{figure}[t]
\centering
  \rotatebox{90}{\resizebox{1.5cm}{!}{\parbox{3cm}{\center\footnotesize \vspace{-7px} Video Object\\Segmentation}}}
  \includegraphics[width=0.23\textwidth, frame]{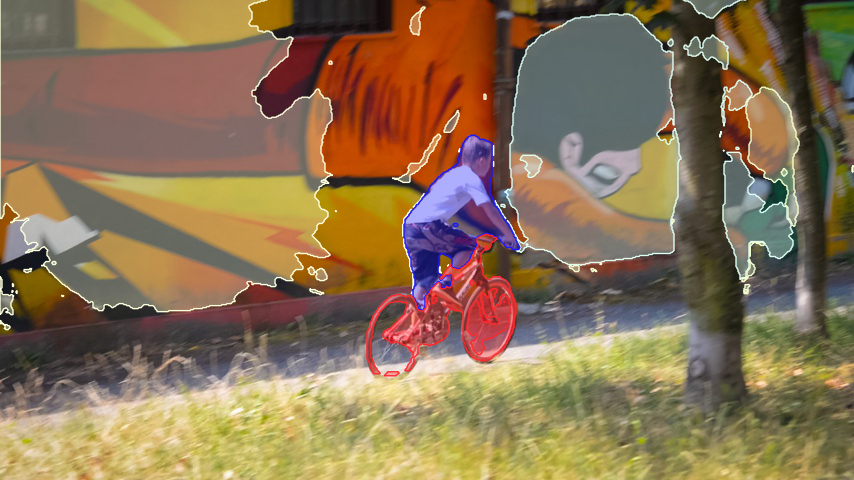}\hspace{1px}%
  \includegraphics[width=0.23\textwidth, frame]{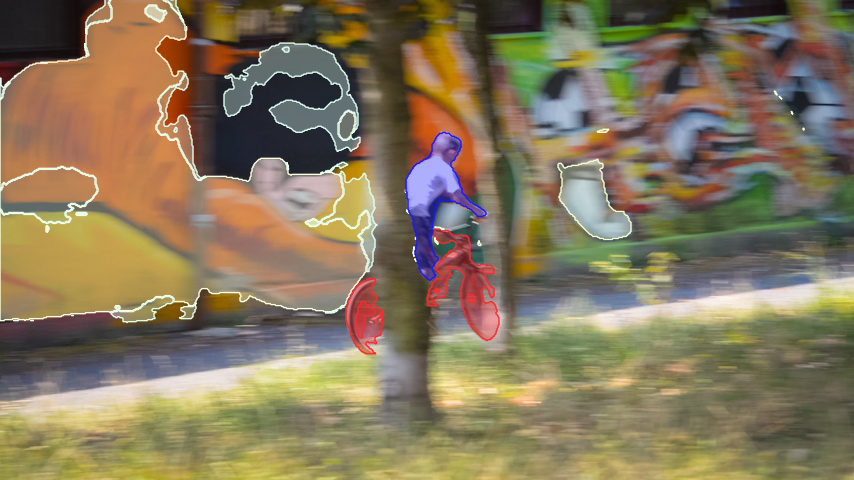}\hspace{1px}%
  \includegraphics[width=0.23\textwidth, frame]{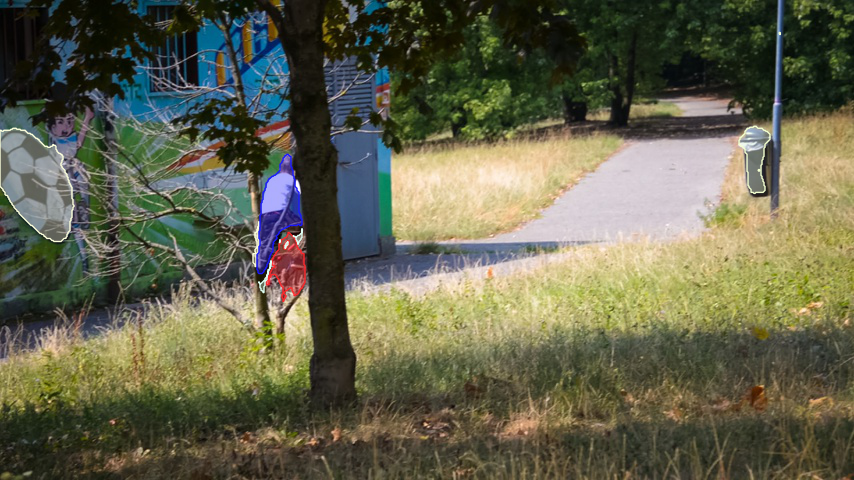}\hspace{1px}%
  \includegraphics[width=0.23\textwidth, frame]{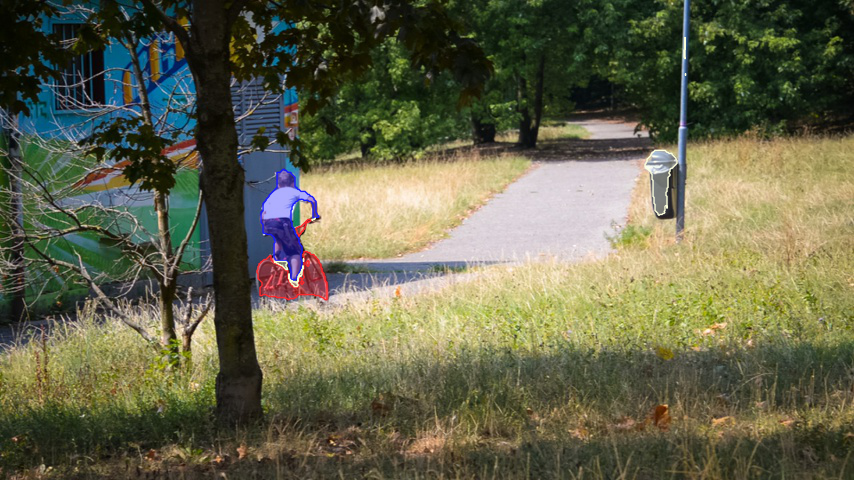}\\
  \rotatebox{90}{\resizebox{1.5cm}{!}{\parbox{3cm}{\center\footnotesize \vspace{-7px} Video Instance\\Segmentation}}}
  \includegraphics[width=0.23\textwidth, trim={330 200 0 0}, clip, frame]{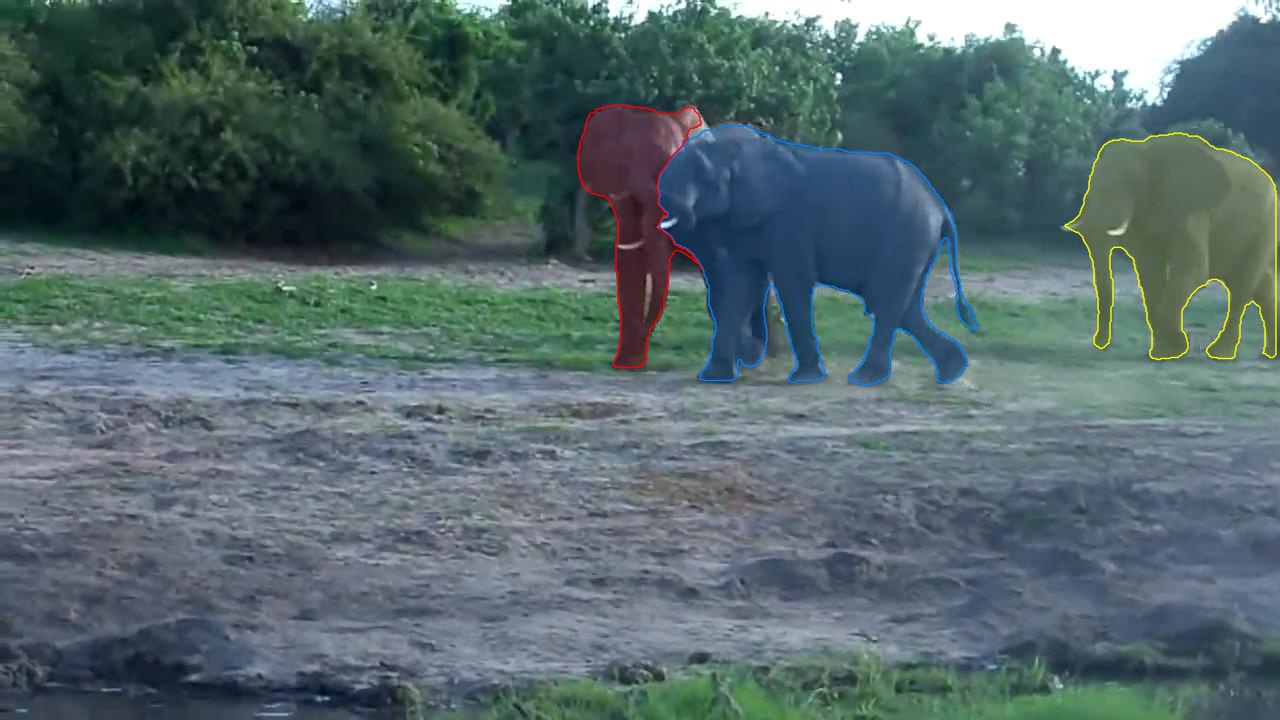}\hspace{1px}%
  \includegraphics[width=0.23\textwidth, trim={330 200 0 0}, clip,frame]{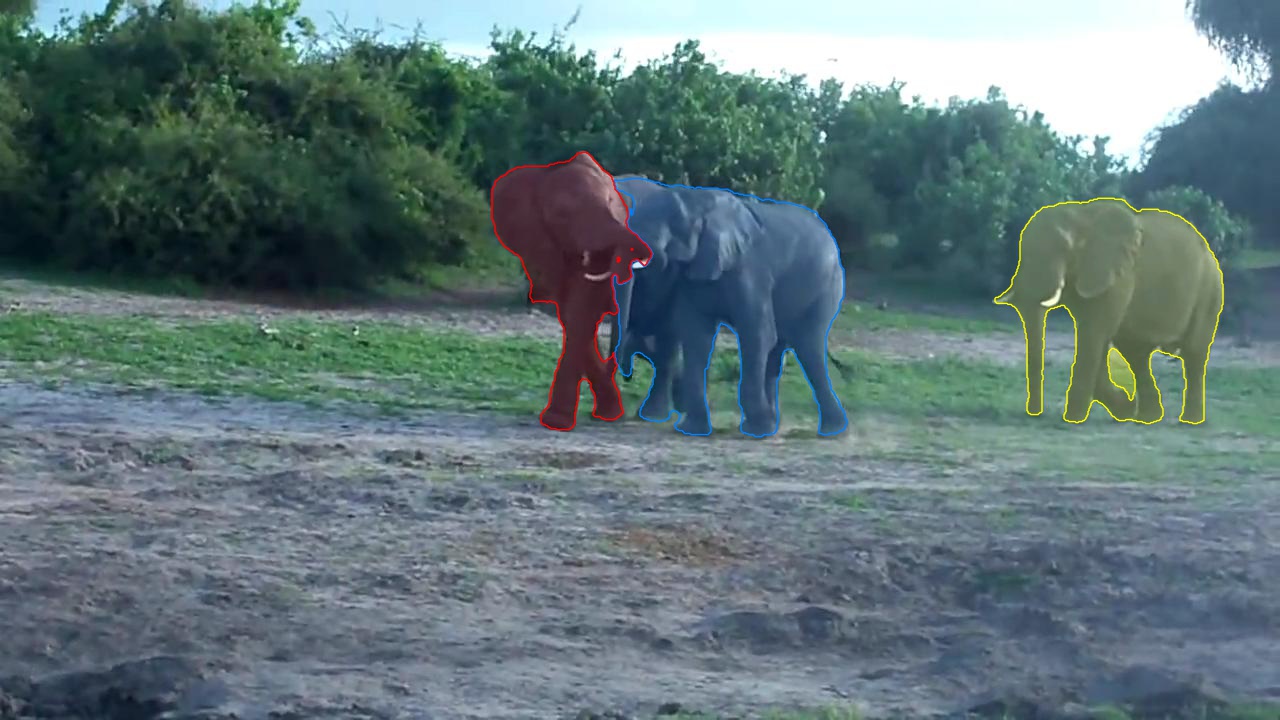}\hspace{1px}%
  \includegraphics[width=0.23\textwidth, trim={330 200 0 0}, clip,frame]{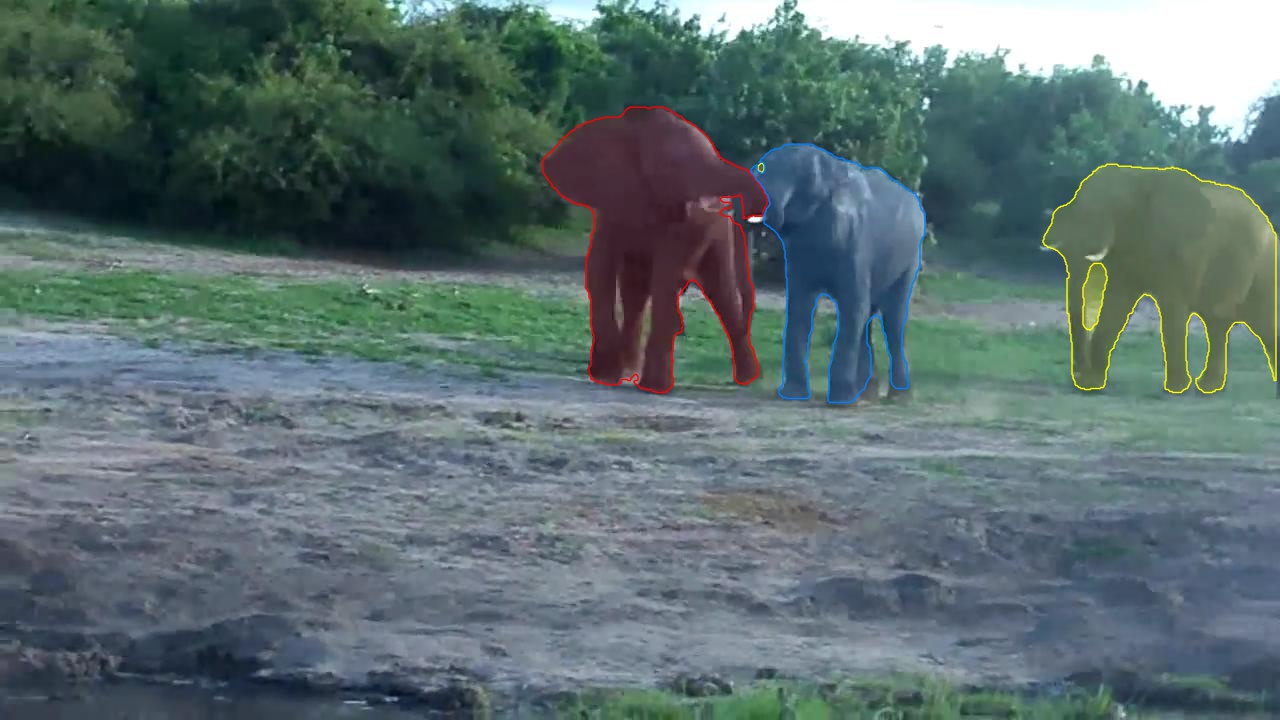}\hspace{1px}%
  \includegraphics[width=0.23\textwidth, trim={330 200 0 0}, clip,frame]{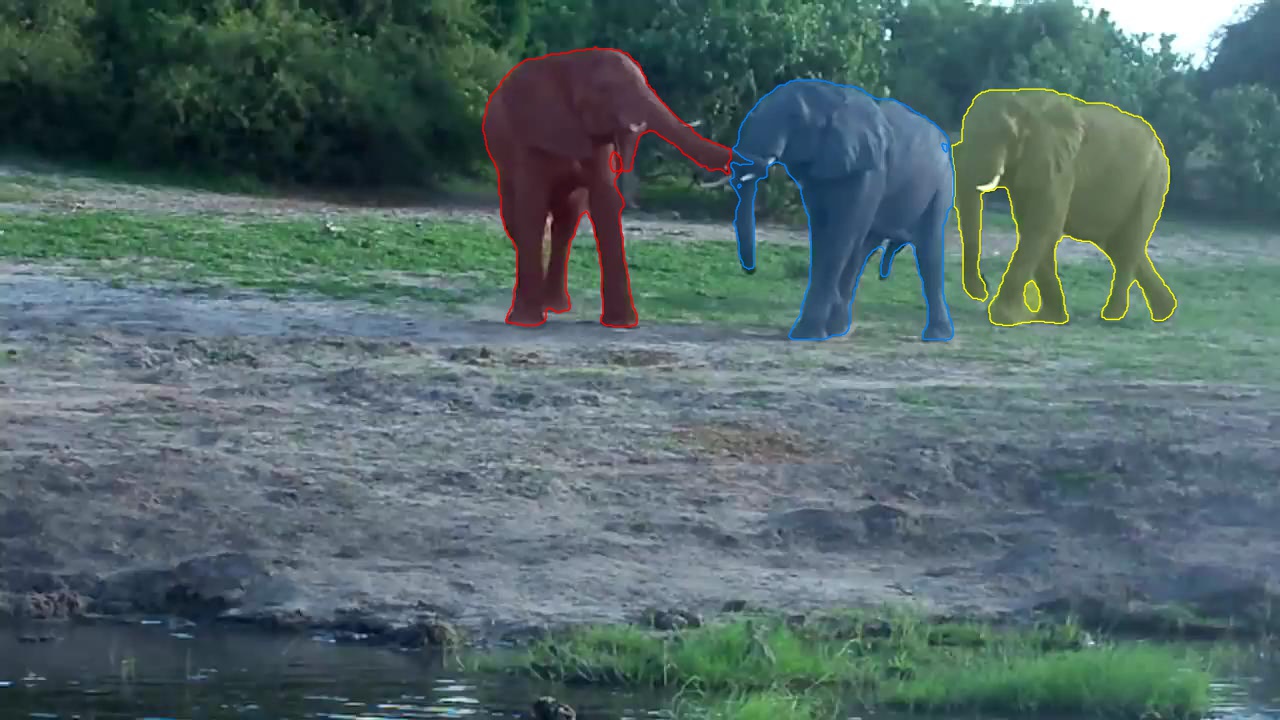}\\
  \rotatebox{90}{\resizebox{1.5cm}{!}{{\parbox{3cm}{\center\footnotesize \vspace{-7px} Multi-Object\\Tracking \& Segment.}}}}
  \includegraphics[width=0.23\textwidth, trim={50 0 520 0}, clip, frame]{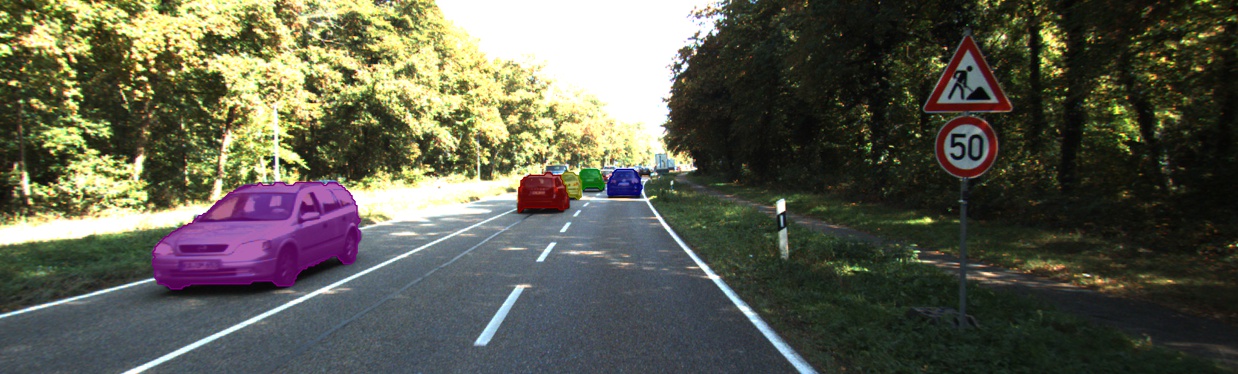}\hspace{1px}%
  \includegraphics[width=0.23\textwidth, trim={50 0 520 0}, clip, frame]{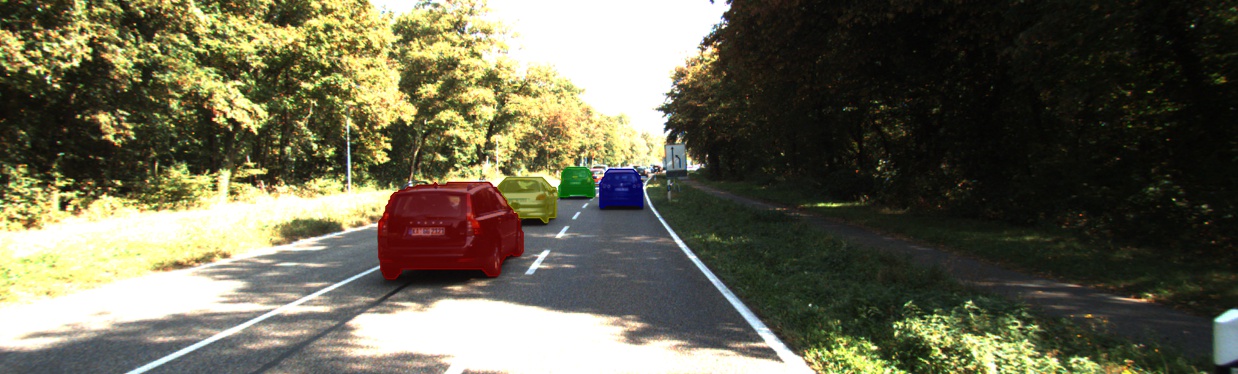}\hspace{1px}%
  \includegraphics[width=0.23\textwidth, trim={50 0 520 0}, clip, frame]{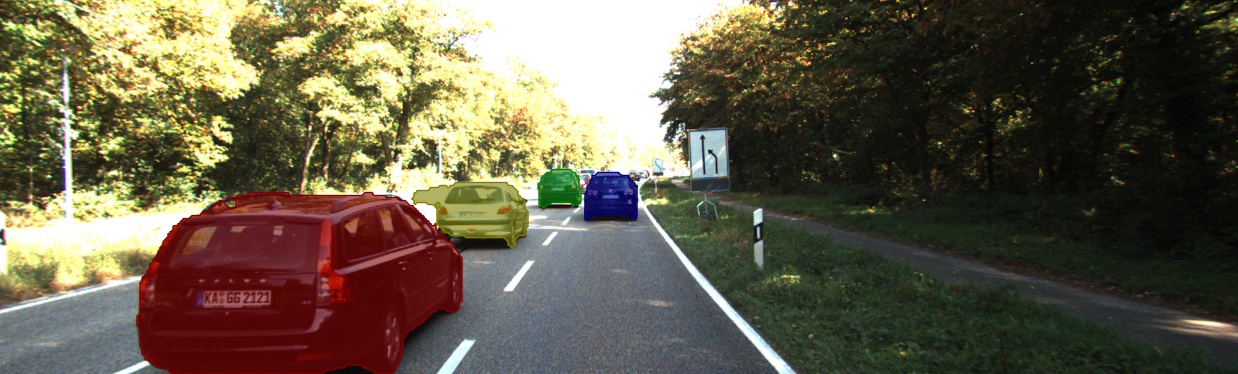}\hspace{1px}%
  \includegraphics[width=0.23\textwidth, trim={50 0 520 0}, clip, frame]{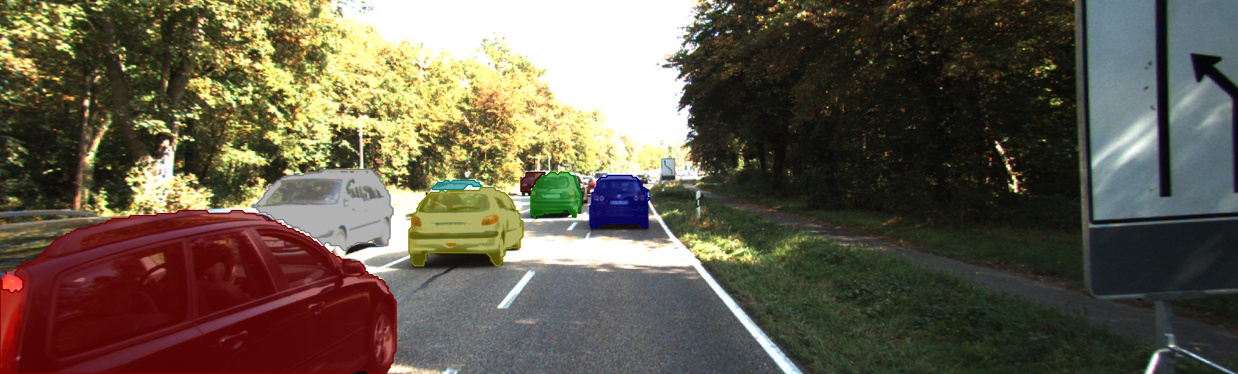}\\
\raggedleft
\begin{tikzpicture}[node distance=2cm]
\node (A) at (2.75, 0) {};
\node (B) at (13.0, 0) {};
\draw[-{Stealth}, to path={-- (\tikztotarget)}](A) edge (B);
\node[text width=1cm] at (13.5,0){time};
\end{tikzpicture}
    \caption{Our method is applicable to multi-object segmentation tasks such as VOS \textit{(top)}, VIS \textit{(middle)} and MOTS \textit{(bottom)}.}
  \label{fig:teaser}
\end{figure}
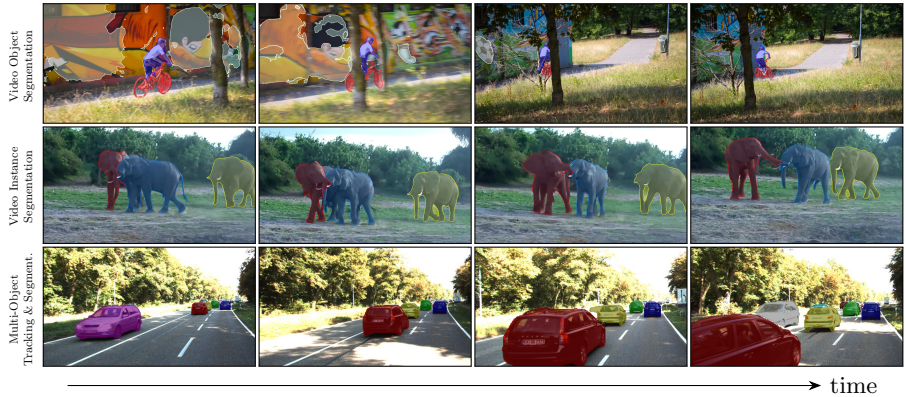

State-of-the-art methods tackling these tasks~\cite{Voigtlaender19CVPR,Wang19ICCVW,Yang19ICCV} usually operate in a \textit{top-down} fashion and follow the \textit{tracking-by-detection} paradigm which is well-established in multi-object tracking (MOT)~\cite{ButtCollins13CVPR,Huang08ECCV,Leibe08TPAMI,Okuma04ECCV}. Such methods usually employ multiple networks to detect objects in individual images~\cite{He17ICCV}, associate them over consecutive frames, and resolve occlusions using learned appearance models.
Though these approaches yield high-quality results, they involve multiple networks, are computationally demanding, and not end-to-end trainable. 

Inspired by the Perceptual Grouping Theory~\cite{Palmer02}, 
% which postulates that the mammalian vision system groups similar looking regions into \textit{object instances}, 
we learn to segment object instances in videos in a \textit{bottom-up} fashion by leveraging \emph{spatio-temporal} embeddings. To this end, 
we propose an efficient, single-stage network that operates directly on a 3D spatio-temporal volume.
We train the embeddings in a category-agnostic setting, such that pixels belonging to the same object instance across the spatio-temporal volume are mapped to a single cluster in the embedding space.
This way, we can infer object instances by simply assigning pixels to their respective clusters.
Our method outperforms proposal-based methods for tasks involving pixel-precise tracking such as Unsupervised Video Object Segmentation (UVOS)~\cite{Caelles17CVPR,Caelles19arXiv}, Video Instance Segmentation (VIS)~\cite{Yang19ICCV}, and Multi-Object Tracking and Segmentation (MOTS)~\cite{Voigtlaender19CVPR}. 

To summarize, our contributions are the following: (i) We propose a unified approach for tasks involving instance segmentation in videos~\cite{Caelles19arXiv,Voigtlaender19CVPR,Yang19ICCV}. Our method performs consistently well under highly varied settings (see \reffig{teaser}) such as automotive driving scenes~\cite{Voigtlaender19CVPR}, semantically diverse YouTube videos~\cite{Yang19ICCV} and scenarios where object classes are not pre-defined~\cite{Caelles19arXiv}.
(ii) We propose using spatio-temporal embeddings for the aforementioned set of tasks. To this end, we propose a set of mixing functions (Sec.~\ref{sec:embedding_representation}) that improve performance by modifying the feature representation of these embeddings. Our method enables a simple inference procedure based on clustering within a 3D spatio-temporal volume, thus alleviating the need for external components for temporal association.
(iii) We propose a single-stage network architecture which is able to effectively incorporate temporal context and learn the spatio-temporal embeddings.

\section{Related Work}

\PAR{Image-level Instance Segmentation:}
Several existing methods for image-level instance segmentation, which is closely related to our task, operate in a \textit{top-down} fashion by using a Region Proposal Network to generate object proposals~\cite{Chen19ICCV,Pinheiro16ECCV,Pinheiro16NIPS,Ren15NIPS}, 
% and then performing classification and pixel-precise segmentation~\cite{He17ICCV}.
which are then classified and segmented~\cite{He17ICCV}.
%
%Such object proposals are common in tracking~\cite{Dave19arxiv,Osep18ICRA,Osep19arxiv,Zulfikar19CVPRW}.
%
% There also exist methods which operate in a \textit{bottom-up} fashion by grouping pixels belonging to the same object instance~\cite{Brabandere17CVPRW,Brabandere17arxiv,Kong18CVPR,Neven19CVPR,Newell17NIPS,Novotny18ECCV}.
Other methods operate in a \textit{bottom-up} fashion by grouping pixels belonging to the same object instance~\cite{Brabandere17CVPRW,Brabandere17arxiv,Kong18CVPR,Neven19CVPR,Newell17NIPS,Novotny18ECCV}.
Novotny~\etal~\cite{Novotny18ECCV} introduce an embedding mixing function to overcome appearance ambiguity, and predict a displacement field from the instance-specific center, which is similar to Hough voting based methods for object detection~\cite{Leibe08IJCV,wang20cvpr}. Recent methods for 3D object detection and instance segmentation also follow this trend~\cite{Engelmann20CVPR,jiang20cvpr,Qi19CVPR,Elich19GCPR,zhang2020spatial}.
Neven~\etal~\cite{Neven19CVPR} extend~\cite{Brabandere17CVPRW} by training a network to predict object centers and clustering bandwidths, thus alleviating the need for density-based clustering algorithms~\cite{Comaniciu02TPAMI,Ester96KDD,Mcinnes2017OSS}. This serves as a basis for our work.
However, in contrast to our approach, the aforementioned methods are only suitable for image-level segmentation. 

\PAR{Video Segmentation:}
Temporally consistent object segmentation in videos can benefit several other tasks such as action/activity recognition~\cite{Gkioxari15CVPR,Hou17ICCV}, video object detection~\cite{Kang17CVPR,Feichtenhofer17ICCV} and object discovery~\cite{Kwak15ICCV,Osep19ICRA,Wang14ECCV,Xiao16CVPR,Dave19arxiv,xie19cvpr}.
Several \textit{bottom-up} methods segment moving regions by grouping based on optical flow~\cite{Van13ICCV,Xu12CVPR} or point trajectories~\cite{Brox10ECCV,Ochs12CVPR,Palou13CVPR,xie19cvpr}. By contrast, our method segments and tracks both moving and static objects.
Other methods obtain video object proposals~\cite{Feichtenhofer17ICCV,Gkioxari15CVPR,Hou17ICCV,Kang17CVPR} based on cues such as actions~\cite{Gkioxari15CVPR,Hou17ICCV} or image-level objects of interest~\cite{Kang17CVPR,Osep19arxiv}. Feature representations are then learned for these proposals in order to perform classification. Instead, we propose a single-stage approach that localizes objects with pixel-level precision.

\PAR{Pixel-precise Tracking of Multiple Objects:}
Multi object tracking (MOT) has its roots in robotic perception~\cite{JainNagel79TPAMI,Paragios00TPAMI,Teichman11ICRA,Wren97TPAMI}. 
Although some works extend MOT with pixel-precise masks~\cite{Milan15CVPR,Osep18ICRA}, a much larger set of works can be found in the domain of Video Object Segmentation (VOS), which encompasses multiple tasks related to pixel-precise tracking. In the \textit{semi-supervised} variant of VOS~\cite{Caelles18arXiv}, the ground-truth masks of the objects which are meant to be tracked are given for the first frame. 
State-of-the-art approaches to this task~\cite{Caelles17CVPR,Ting17NIPS,Oh2019ICCV,Tokmakov17ICCV,Ventura19CVPR,Voigtlaender19CVPRFeelVOS,Wug18CVPR,Xu18ECCV,Yang18CVPR,Zeng19ICCV} usually involve online fine-tuning on the first frame masks and/or 
follow the \textit{object proposal generation and mask propagation} approach~\cite{Luiten18ACCV,Zeng19ICCV}.
Chen~\etal~\cite{Chen2018CVPR} tackle this problem by learning embeddings from the first-frame masks, and then associating pixels in the remaining frames. % to these learned embeddings.

More relevant to our work is the task of \textit{unsupervised} VOS~\cite{Caelles19arXiv,Perazzi16CVPR}. 
Here, no ground truth information is provided at test-time, and the goal is to segment and track all dominant ``foreground" objects in the video clip. 
Current state-of-the-art methods for this task are either proposal-based~\cite{Zeng19ICCV,Zulfikar19CVPRW} or focus on foreground-background segmentation~\cite{Hou19BMVC,Jain17CVPR,SiamICRA2019,Song18ECCV,Tokmakov17ICCV,Ventura19CVPR,Yang19ICCVAnchorDiff,Koh17CVPR}. Li~\etal~\cite{Li18CVPR} propose an approach that groups pixel embeddings based on \textit{objectness} and optical flow cues. 
In contrast to ours, this method processes each frame separately, employs K-means clustering for object localization, and cannot separate different object instances. 
Wang~\etal~\cite{Wang19ICCV} employ an attentive graph neural network~\cite{Gori05IJCNN} and use differentiable message passing to propagate information across time; we compare our results to theirs in \refsec{saliency_seg}.

% -------------------MOTS--------------------------
Recently, the task of Multi-Object Tracking and Segmentation (MOTS)~\cite{Voigtlaender19CVPR} and Video Instance Segmentation (VIS)~\cite{Yang19ICCV} were introduced. 
Voigtlaender~\etal~\cite{Voigtlaender19CVPR} extended the KITTI~\cite{Geiger12CVPR} and MOTChallenge~\cite{LealTaixe15arxiv,Milan16arxiv} datasets with pixel-precise annotations, and proposed a baseline method that adapts Mask R-CNN~\cite{He17ICCV} to associate object proposals across time. Hu~\etal~\cite{Hu19Arxiv} use a mask network to filter foreground embeddings and perform mean-shift clustering to generate object masks, which are then associated over time using a distance threshold unlike our single-stage end-to-end method. 

The YouTube-VIS~\cite{Yang19ICCV} contains a large number of YouTube videos with per-pixel annotations for a variety of object classes. Compared to MOTS, these videos contain fewer instances, but are significantly more diverse. 
In addition to adapting several existing methods~\cite{Bochinski17AVSS,Han16arxiv,Voigtlaender19CVPRFeelVOS,Wojke17ICIP,Yang18CVPR} for this task, Yang ~\etal~\cite{Yang19ICCV} also proposed their own method (MaskTrack-RCNN) which extends Mask R-CNN with additional cues for data association. 
Methods such as~\cite{Dong19ICCVW,Feng19ICCVW,Liu19ICCVW} also rely on object proposals and/or heuristic post-processing to associate objects over time, unlike our end-to-end bottom-up approach.

\section{Our Method}

We tackle the task of instance segmentation in videos by modeling the video clip as a 3D spatio-temporal volume and using a network to learn an embedding for each pixel in that volume. This network is trained to push pixels belonging to different object instances towards different, non-overlapping clusters in the embedding space. 
This differs from most existing approaches, which first generate object detections per-frame, and then associate them over time. The following sections explain our problem formulation, the network architecture and loss functions employed, and the inference process.

\subsection{Problem Formulation}
\label{sec:problem_formulation}

As input, we assume a video clip with $T$ frames of resolution $H\times W$. Let us denote the set of RGB pixels in this clip with $\mathcal{X} \in \mathbb{R}^{N\times 3}$ where $N=T\times H\times W$. 
Assuming there are $K$ object instances in this clip ($K$ is unknown), our aim is to produce a segmentation that assigns each pixel in the clip to either the background, or to exactly one of these $K$ instances.
We design a network that predicts video instance tubes by clustering pixels simultaneously across space and time based on a learned embedding function. 
Instead of learning just the embedding function and then using standard density-based clustering algorithms~\cite{Fukunaga1975TOIT,Lloyd1982TOIT,Mcinnes2017OSS} to obtain instance tubes, we take inspiration from Neven~\etal~\cite{Neven19CVPR} and design a network which estimates the cluster centers as well as their corresponding variances, thus enabling efficient inference.
Formally, our network can be viewed as a function that maps the set of pixels $\mathcal{X}$ to three outputs; (1) $\mathcal{E} \in \mathbb{R}^{N\times E}$: an $E$-dimensional embedding for each pixel, (2) $\mathcal{V} \in \mathbb{R}_+^{N\times E}$: a positive variance for each pixel and each embedding dimension, and (3) $\mathcal{H} \in [0,1]^N$: an object instance center heat-map.

\PAR{Instance representation:} The network is trained such that the embeddings belonging to the $j^\text{th}$ instance in the video clip are modeled by a multivariate Gaussian distribution $\mathcal{N}(\vec{\mu}_j,\matrixvar{\Sigma}_j)$. 
Assuming that this instance comprises the set of pixel coordinates $\mathcal{C}_j$ with cardinality $N_j$, we denote the embeddings and variances output by the network at these coordinates using 
$\mathcal{E}_j \subset \mathcal{E}$, $\mathcal{E}_j \in \mathbb{R}^{N_j \times E}$ 
and 
$\mathcal{V}_j \subset \mathcal{V}$, $\mathcal{V}_j \in \mathbb{R}_+^{N_j \times E}$, respectively.
During training, $\mathcal{C}_j$ (\ie, the ground truth mask tube for instance $j$) is known. Using it, the mean $\vec{\mu}_j$ and covariance $\matrixvar{\Sigma}_j$ of the distribution are computed by averaging over the per-pixel outputs:

\begin{equation}
    \vec{\mu}_j = \frac{1}{N_j}\sum\limits_{\vec{e} \hspace{1pt} \in \hspace{1pt} \mathcal{E}_j} \vec{e} \in \mathbb{R}^E, \hspace{20pt} \matrixvar{\Sigma}_j = \frac{1}{N_j}\hspace{2pt} \text{diag}\left(\sum\limits_{\vec{v} \hspace{1pt}\in \hspace{1pt} \mathcal{V}_j} \vec{v}\right) \in \mathbb{R}^{E\times E}~.
\label{eq:mean_var}
\end{equation}
This single distribution models all embeddings belonging to instance $j$ across the entire clip (not individually for each frame).
We can now use the distribution $\mathcal{N}(\vec{\mu}_j,\matrixvar{\Sigma}_j)$ to compute the probability $p_{ij}$ of any embedding $\vec{e}_i \in \mathcal{E}$, anywhere in the input clip, of belonging to instance $j$:
\begin{equation}
    p_{ij} = \frac{1}{{(2\pi)}^\frac{E}{2}|\boldsymbol{\Sigma_j}|^\frac{1}{2}}\text{exp}\left(-\frac{1}{2}(\vec{e}_i - \vec{\mu}_j)^T \matrixvar{\Sigma}_j^{-1}(\vec{e}_i - \vec{\mu}_j)\right)~.
\label{eq:probs}
\end{equation}
Using \refequ{probs} to compute $p_{ij}$, $\forall\hspace{2pt} i \in \{1, ..\hspace{1pt} , N\}$, we can obtain the set of pixels $\widehat{\mathcal{C}}_j$ comprising the predicted mask tube for instance $j$ by thresholding the probabilities at 0.5:

\begin{equation}
    \widehat{\mathcal{C}}_j = \{\hspace{1pt}  (x_i, y_i, t_i) \hspace{2pt} | \hspace{2pt} i \in \{1, ..\hspace{1pt} , N\} ,\hspace{3pt} p_{ij} > 0.5 \hspace{1pt}\}~.
\label{eq:thresh}
\end{equation}
\PAR{Training:} This way, the training objective can be formulated as one of learning the optimal parameters $\vec{\mu}_j^\text{opt}$ and $\matrixvar{\Sigma}_j^\text{opt}$ which maximize the intersection-over-union (IoU) between the predicted and ground-truth mask tubes in the clip: 

\begin{equation}
    \vec{\mu}_j^\text{opt}, \matrixvar{\Sigma}_j^\text{opt} = \underset{\vec{\mu}_j, \matrixvar{\Sigma}_j}{\text{argmax}} \hspace{5pt}\frac{\widehat{\mathcal{C}}_j \cap \mathcal{C}_j}{\widehat{\mathcal{C}}_j \cup \mathcal{C}_j},
\label{eq:optimization}
\end{equation}
that is, all pixels in the ground truth mask tube $\mathcal{C}_j$ should have probability larger than $0.5$, and vice versa. A classification loss such as cross-entropy could be used to optimize the pixel probabilities. However, we employ the Lov\`{a}sz hinge loss~\cite{Berman18CVPR,Berman18CVPR2,Neven19CVPR,Yu2015ICML} (details in supplementary material), which is a differentiable, convex surrogate of the Jaccard index that directly optimizes the IoU. This formulation allows $\vec{\mu}_j$ and $\matrixvar{\Sigma}_j$ to be implicitly learned by the network. 

Using Eqs.~\ref{eq:mean_var}-\ref{eq:optimization}, we can define and optimize the distribution for every instance (\ie $\forall{j} \in 
\{1,...,K\}$). Note that only a single forward pass of the network is required regardless of the number of instances in the clip.
This is in contrast to common approaches for Video Object Segmentation, which only process one instance at a time~\cite{Oh2019ICCV,Wug2018CVPR}.

We further remark that ours is a \textit{bottom-up approach} which detects and tracks objects in a single step, thus mitigating the inherent drawback of \textit{top-down} approaches that often require different networks/cues for single-image object detection and temporal association. A further advantage of our approach is that it can implicitly resolve occlusions, insofar as they occur within the clip. 

\subsection{Embedding Representation}
\label{sec:embedding_representation}

Under the formulation described in \refsec{problem_formulation}, the network can learn arbitrary representations for the embedding dimensions. However, it is also possible to fix the representation by using a mixing function $\phi: \mathbb{R}^E \rightarrow \mathbb{R}^E$ that modifies the embeddings $\mathcal{E}$ as follows: $\mathcal{E} \gets \{\hspace{1pt} \phi(\vec{e}),\hspace{2pt} |\hspace{2pt} \vec{e} \in \mathcal{E} \hspace{1pt}\}$.

In~\cite{Neven19CVPR}, for the task of single-image instance segmentation, 2D embeddings were used ($E=2$) in conjunction with a spatial coordinate mixing function $\phi_\text{xy}(\vec{e}_i) = \vec{e}_i + [x_i, y_i]$
\footnote{This notation denotes element-wise addition between $\boldsymbol{e}_i$ and $[x_i, y_i]$}.
With this setting, the embeddings could be interpreted as offsets to the $(x,y)$ coordinates of their respective locations. The network thus learned to cluster the embeddings belonging to a given object towards some object-specific point on the image. It follows that the predicted variances could be interpreted as the network's estimate of the size of the object along the $x$ and $y$ axes. We postulate that the reason for this formulation yielding good results is that the $(x,y)$ coordinates already serve as a good initial feature for instance separation; the network can then enhance this representation by producing offsets which further improve the clustering behavior. Furthermore, this can be done in an end-to-end trainable fashion which allows the network to adjust the clustering parameters, \ie, the Gaussian distribution parameters, for each instance. In general, it has been shown that imparting spatial coordinate information to CNNs can improve performance for a variety of tasks~\cite{Liu2018NIPS,Novotny18ECCV}.

Compared to single-image segmentation, the task of associating pixels across space and time in in videos poses additional challenges, \eg, camera ego-motion, occlusions, appearance/pose changes. To tackle these challenges we propose (and experimentally validate in \refsec{ablation}) the following extensions:

\PAR{Spatio-temporal coordinating mixing:} Since we operate on video clips instead of single images, a logical extension is to use 3D embeddings ($E$\,=\,$3$) with a spatio-temporal coordinate mixing function $\phi_\text{xyt}(\vec{e}_i) = \vec{e}_i + [x_i, y_i, t_i]$.

\PAR{Free dimensions:} In addition to the spatial (and temporal) coordinate dimensions, it can be beneficial to include extra dimensions whose representation is left for the network to decide. The motivation here is to improve instance clustering quality by allowing additional degrees of freedom in the embedding space. From here on, we shall refer to these extra embedding dimensions as \textit{free dimensions}. For example, if $E=4$ with 2 spatial coordinate dimensions and 2 free dimensions, the mixing function is denoted as $\phi_\text{xyff}(\vec{e}_i) = \vec{e}_i + [x_i, y_i, 0, 0]$. 

There is, however, a caveat with free dimensions: since the spatial (and temporal) dimensions already achieve reasonable instance separation, the network may converge to a poor local minimum by producing very large variances for the free dimensions instead of learning a discriminative feature representation. Consequently, the free dimensions may end up offering no useful instance separation during inference. We circumvent this problem at the cost of introducing one extra hyper-parameter by setting the variances for the free dimensions to a fixed value $\vec{v}_\text{free}$.
We justify our formulation quantitatively using multiple datasets and different variants of the mixing function $\phi(\cdot)$ in \refsec{ablation}.

\begin{figure}[t]
    \centering
    \includegraphics[width=0.9\linewidth,height=4cm]{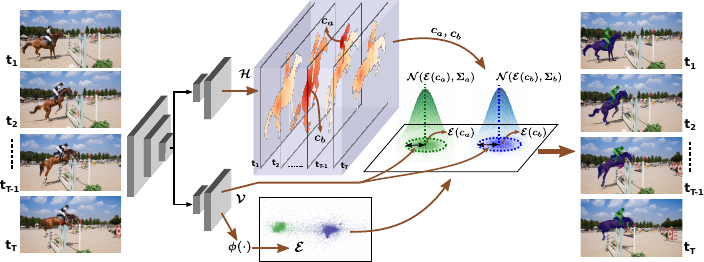}
    \caption{For an input video clip, our network produces embeddings ($\mathcal{E}$), variances ($\mathcal{V}$), and instance center heat map ($\mathcal{H}$). $\mathcal{H}$ contains one peak per object for the entire spatio-temporal volume ($c_a$ for the rider, $c_b$ for the horse). $\mathcal{E}(c_a)$, $\mathcal{E}(c_b)$ and $\mathcal{V}(c_a)$, $\mathcal{V}(c_b)$ are the corresponding embeddings and variances at $c_a$ and $c_b$, respectively. These quantities are then used to define the Gaussian distribution for each object.}
    \label{fig:pipeline}
\end{figure}

\subsection{Inference}
\label{sec:inference}

Since the ground truth mask tube is not known during inference, it is not possible to obtain $\vec{\mu}_j$ and $\matrixvar{\Sigma}_j$ using  \refequ{mean_var}. This is where the instance center heat map $\mathcal{H}$ comes into play. For each pixel $\vec{c}_i = (x_i, y_i, t_i)$, the value $\mathcal{H}(\vec{c}_i) \in [0, 1]$ in the heat map at this location gives us the probability of the embedding vector $\mathcal{E}(\vec{c}_i)$ at this location being an instance center. The sequential process of inferring object instances in a video clip is described in the following algorithm:

\begin{enumerate}
    \item Identify the coordinates of the most likely instance center $\vec{c}_j = {\text{argmax}_i} \hspace{3pt}\mathcal{H}(\vec{c}_i)$.
    \item Find the corresponding embedding vector $\mathcal{E}(\vec{c}_j)$ and variances $\mathcal{V}(\vec{c}_j)$.
    \item Using $\vec{\mu}_j \gets \mathcal{E}(\vec{c}_j)$ and $\matrixvar{\Sigma}_j \gets \text{diag}\hspace{2pt}(\mathcal{V}(\vec{c}_j))$, generate the 3D mask tube $\widehat{\mathcal{C}}_j$ for this instance by computing per-pixel probabilities using \refequ{probs}, and then thresholding them as in \refequ{thresh}.
    \item Since the pixels in $\widehat{\mathcal{C}}_j$ have now been assigned to an instance, the embeddings, variances and heat map probabilities at these pixel locations are masked out and removed from further consideration:
    \begin{equation}
        \mathcal{E} \gets \mathcal{E} \setminus \widehat{\mathcal{E}}_j \hspace{2pt}, \hspace{20pt} \mathcal{V} \gets \mathcal{V} \setminus \widehat{\mathcal{V}}_j \hspace{2pt}, \hspace{20pt} \mathcal{H} \gets \mathcal{H} \setminus \widehat{\mathcal{H}}_j.
    \end{equation}
    \item Repeat steps 1-4 until either $\mathcal{E} = \mathcal{V} = \mathcal{H} = \emptyset$, or the next highest probability in the heat map falls below some threshold.
\end{enumerate}

Even though this final clustering step (\reffig{pipeline}) depends on the number of instances in a clip, in practice, the application of \refequ{probs} and~\ref{eq:thresh} carries little computational overhead and its run-time is negligible compared to a forward pass.

\PAR{Video clip stitching:} Due to memory constraints, the clip length that can be input to the network is limited. In order to apply our framework to videos of arbitrary length, we split the input video into clips of length $T$ with an overlap of $T_\text{c}$ frames between consecutive clips. Linear assignment~\cite{Kuhn55NRLQ} is then used to associate the predicted tracklets in consecutive clips. The cost metric for this assignment is the IoU between tracks in overlapping frames. Our approach is currently \textit{near online} because, given a new frame, the delay until its output becomes available is at most $T-T_\text{c}-1$.

\subsection{Losses}
Our model's loss function is a linear combination of three  terms:
\begin{equation}
    L_\text{total} = L_\text{emb} + L_\text{smooth} + L_\text{center}
\end{equation}
\PAR{Embedding loss $L_\text{emb}$:} As mentioned in  \refsec{problem_formulation}, we use the Lov\`{a}sz hinge loss to optimize the IoU between the predicted and ground truth masks for a given instance. The embedding loss for the entire input clip is calculated as the mean of the Lov\`{a}sz hinge loss for all object instances in that clip.

\PAR{Variance smoothness loss $L_\text{smooth}$:} To ensure that the variance values at every pixel belonging to an object are consistent, we employ a smoothness loss $L_\text{smooth}$ similar to ~\cite{Neven19CVPR}. This regresses the variances $\mathcal{V}_j$ for instance $j$ to be close to the average value of all the variances for that instance, \ie $\mathrm{Var}[\mathcal{V}_j]$.

\PAR{Instance center heat map loss $L_\text{center}$:} For all pixels belonging to instance $j$, the corresponding outputs in the sigmoid activated heat map $\mathcal{H}_j$ are trained with an $L_\text{2}$ regression loss to match the output of \refequ{probs}. The outputs for background pixels are regressed to 0. During inference, this enables us to sample the highest values from the heat map which corresponds to the peak of the learned Gaussian distributions for the object instances in an input volume, as explained in Sec~\ref{sec:inference}.

\begin{figure}[t]
    \centering
    \includegraphics[width=0.9\linewidth,height=4.5cm]{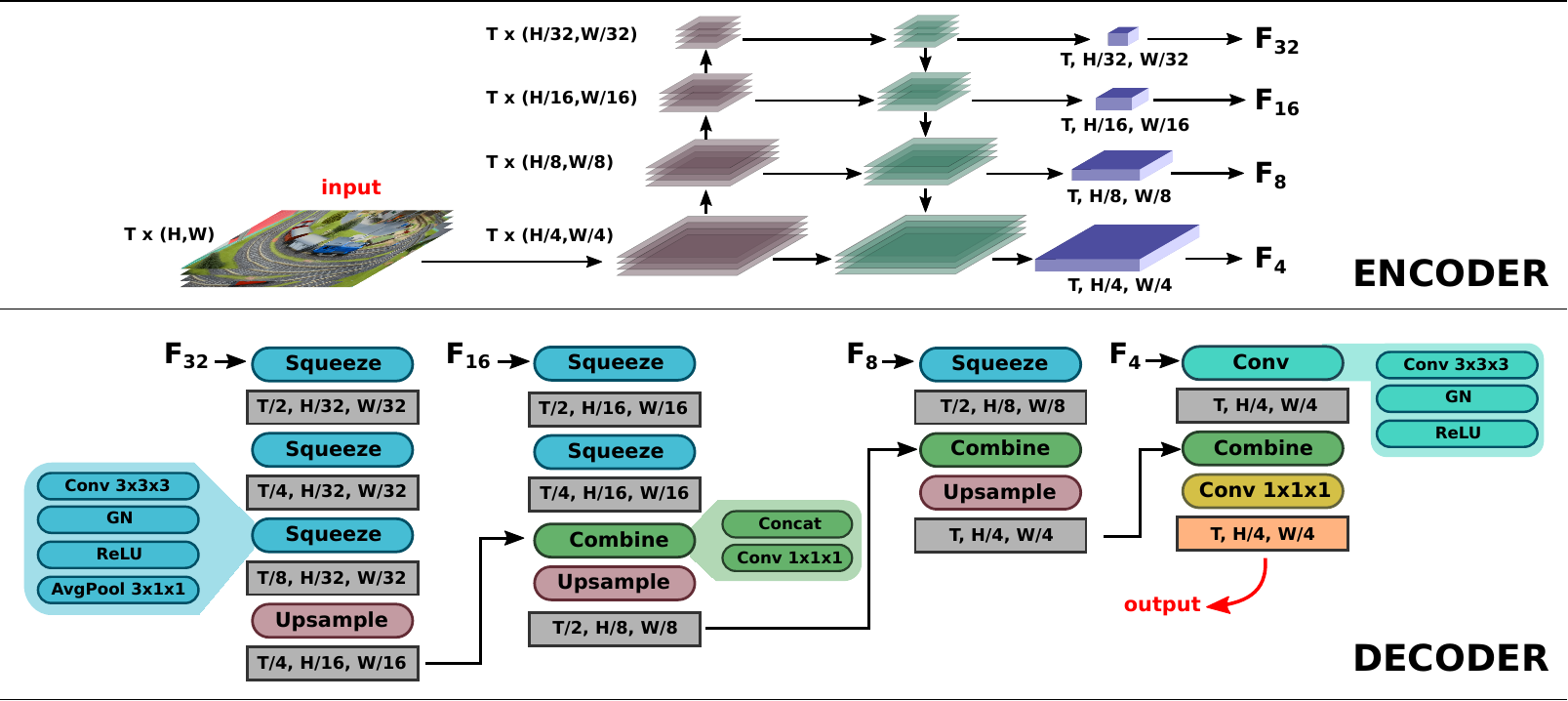}
    \caption{The network has an encoder-decoder structure. GN: Group Normalization~\cite{Wu18ECCV}.}
    \label{fig:network_architecture}
\end{figure}

\subsection{Network Architecture}
\label{sec:network_architecture}

The network (\reffig{network_architecture}) consists of an encoder with two decoders. The first decoder outputs the embeddings $\mathcal{E}$ and variances $\mathcal{V}$, while the second outputs the instance center heat map $\mathcal{H}$. The encoder comprises a backbone with Feature Pyramid Network (FPN) that produces feature maps at 4 different scales for each image in the input clip. The feature maps at each scale are then stacked along the temporal dimension before being input to each of the decoders. 

Our decoder consists of 3D convolutions and pooling layers which first compress the feature maps along the temporal dimension before gradually expanding them back to the input size. The underlying idea here is to allow the network to learn temporal context in order to enable pixel association across space and time. To reduce the decoder's memory footprint and run-time, the large sized feature maps undergo a lower degree of temporal pooling (\ie, fewer convolution/normalization/pooling layers). We call our decoder a \textit{temporal squeeze-expand decoder} (abbreviated as TSE decoder). In \refsec{saliency_seg}, we experimentally validate our network's stand-alone ability to learn spatio-temporal context.

\subsection{Category Prediction}
\label{sec:category_classification}

Our task formulation is inherently category-agnostic. However, some datasets~\cite{Voigtlaender19CVPR,Yang19ICCV} require a category label for each predicted object track. For such cases we add an additional TSE decoder to the network that performs semantic segmentation for all pixels in the input clip, and that is trained using a standard cross-entropy loss. During inference, the logits for all pixels belonging to a given object instance are averaged, and the highest scoring category label is assigned. Note that this is merely a post-processing step; the instance clustering still happens in a category-agnostic manner.

\section{Experimental Evaluation}

To demonstrate our method's effectiveness and generalization capability, we apply it to three different tasks and datasets involving pixel-precise segmentation and tracking of multiple objects in videos. 

\subsection{Training}
\label{sec:training}
For all experiments, we use a ResNet-101\,\cite{He16CVPR} backbone initialized with weights from 
Mask R-CNN\,\cite{He17ICCV} trained for image instance segmentation on COCO\,\cite{Lin14ECCV}. The temporal squeeze-expand decoders are initialized with random weights. The network is optimized end-to-end using SGD with momentum 0.9 and an initial learning rate of $10^{-3}$ which decays exponentially.

\PAR{Augmented Images.} 
Since the amount of publicly available video data with dense per-pixel annotations is limited, we utilize image instance segmentation datasets by synthesizing training clips from single images. We apply on-the-fly random affine transformations and motion blur and show that this technique is effective (\refsec{ablation}) even though such augmented image sequences have little visual resemblance to video clips.

\subsection{Benchmarks}
\label{sec:benchmarks}

\PARbegin{DAVIS Unsupervised:} DAVIS\,\cite{Caelles19arXiv} is a popular video object segmentation dataset with 90 videos (60 for training and 30 for validation) containing multiple moving objects of diverse categories. Several DAVIS benchmarks have been introduced over the years; we evaluate our method on the 2019 \textit{Unsupervised Video Object Segmentation} (UVOS) benchmark in which the salient ``foreground objects'' in each video have to be segmented and tracked.
The evaluation measures employed are (i) $\mathcal{J}$-score (the IoU between predicted and ground truth mask tubes), and (ii) $\mathcal{F}$-score (accuracy of predicted mask boundaries against ground truth). The mean of those measures, $\mathcal{J}\text{\&}\mathcal{F}$, serves as the final score.

\PAR{YouTube-VIS:} The YouTube Video Instance Segmentation (YT-VIS)\,\cite{Yang19ICCV} dataset contains $2,\!883$ high quality YouTube videos with $\sim$131k object instances spanning $40$ known categories. 
The task requires all objects belonging to the known category set to be segmented, tracked and assigned a category label.
The evaluation measures used for this task are an adaptation of the Average Precision ($\mathcal{AP}$) and Average Recall ($\mathcal{AR}$) metrics used for image instance segmentation. 
\PAR{KITTI-MOTS:} Multi-Object Tracking and Segmentation (MOTS) \cite{Voigtlaender19CVPR} extends the KITTI multi-object tracking dataset~\cite{Geiger12CVPR} with pixel-precise instance masks. 
It contains 21 videos (12 for training and 9 for validation) of driving scenes captured from a moving vehicle. The task here is to segment and track all \textit{car} and \textit{pedestrian} instances in the videos. The evaluation measures used are an extension of the CLEAR MOT measures\,\cite{Bernardin08JIVP} to account for pixel-precise tracking (details in \cite{Voigtlaender19CVPR}). We report these measures separately for each object class.

\subsection{Ablation Studies}

\PAR{Embedding formulation:}
We first ablate the impact of using different mixing functions for the embeddings (\refsec{embedding_representation}) on the DAVIS'19 \texttt{val} dataset with clip length $T$\,=\,$8$, and summarize the results in \reftab{ablation_embeddings}. 
Compared to the identity function baseline $\phi_{\text{identity}}$, imparting a spatial coordinate offset ($\phi_\text{xy}$) improves the $\mathcal{J}\text{\&}\mathcal{F}$ from $57.3\%$ to $61.6\%$. 
Adding another embedding dimension with a temporal coordinate offset, as in $\phi_\text{xyt}$, yields a further improvement to $62.6\%$. 

Comparing the mixing function pairs ($\phi_\text{xyf}$, $\phi_\text{xyt}$) where $E=3$, and ($\phi_\text{xyff}$, $\phi_\text{xytf}$) where $E=4$, we note that having a free dimension is slightly better than having a temporal dimension since there is a difference of $0.2\%$ $\mathcal{J}\text{\&}\mathcal{F}$ for both pairs of functions. 
This is the case for both DAVIS'19 and YT-VIS,\footnote{\label{fnote:embedding_formulation}results for various $\phi(\cdot)$ on YT-VIS and KITTI-MOTS are given in supplementary} where $\phi_\text{xyff}$ yields the best results. For KITTI-MOTS however,\footnotemark[\value{footnote}] temporal and free dimensions yield roughly the same performance. Our intuitive explanation is that for DAVIS and YT-VIS, object instances normally persist throughout the video clip. 
Therefore, in contrast to the spatial coordinates which serve as a useful feature for instance separation, the temporal coordinate provides no useful separation cue, thus rendering the temporal embedding dimension less effective. 
On the other hand, using a free dimension enables the network to learn a more discriminative feature representation that can better aid in separating instances. 
By contrast, objects in KITTI-MOTS driving scenes undergo fast motion and often enter/exit the scene midway through a clip. Thus, the temporal dimension becomes useful for instance separation.

Having additional embedding dimensions is beneficial, but only up to a certain point (\ie, $E=4$). 
Beyond that, test-time performance drops, as can be seen by comparing $\phi_\text{xyff}$ and $\phi_\text{xyfff}$. 
We conclude by noting that our proposed formulations for $\phi(\cdot)$ improve performance on video-related tasks compared to existing formulations for single-image tasks\,\cite{Neven19CVPR,Novotny18ECCV}. For further discussion and results we refer to the supplementary material.

\label{sec:ablation}
\begin{table}[t]
    \centering
    \footnotesize
    \begin{subtable}[t]{0.25\linewidth}
    {%\captionsetup[table]{skip=0pt}
    \vspace{-2mm}
                \begin{tabular}[t]{p{2.5cm} | c | c }
\toprule
             {\footnotesize Mixing Function} & $\mathcal{J}$\&$\mathcal{F}$ & $E$ \\
            \cmidrule(lr){1-3}
                            $\phi_{\text{identity}}$ & 57.3  & 2\\
                            $\phi_{\text{xy}}$  & 61.6       & 2\\
                            $\phi_\text{xyt}$   & 62.6       & 3\\
                            $\phi_\text{xyf}$   & 62.8       & 3\\
                            $\phi_\text{xytf}$  & 64.2       & 4\\
                            $\phi_\text{xyff}$   & 64.4      & 4\\
                            $\phi_\text{xyfff}$  & 62.4      & 5\\
            \bottomrule
            \end{tabular}
            \caption{}
            \label{tab:ablation_embeddings}
        }
        \end{subtable}  
        \hfill
        \hspace{5mm}
        \begin{subtable}[t]{0.25\linewidth}
        {
        \vspace{-2mm}
        \begin{tabular}[t]{c | c}
            \toprule
            Clip Length (\textit{T}) & $\mathcal{J}\&\mathcal{F}$ \\
            % \cmidrule(lr){1-2}
            \cmidrule(lr){1-2}
                % $\text{T=4 Frames}$  & 62.2 \\
                % $\text{T=8 Frames}$   & 64.4\\
                % $\text{T=16 Frames }$  & 64.7\\
                % $\text{T=24 Frames }$  & 63.1\\
                $\text{4}$  & 62.2 \\
                $\text{8}$   & 64.4\\
                $\text{16}$  & 64.7\\
                $\text{24}$  & 63.1\\
                \\
                \\
                \\
            \bottomrule
            \end{tabular}
            \caption{}
            \label{tab:ablation_temporal}
        }
        \end{subtable}%
        \hfill
        \begin{subtable}[t]{0.36\linewidth}%
        {
        % \vspace{-2mm}
            \begin{tabular}[t]{lr}%
            {\scriptsize\textbf{(c)}}&%
            \begin{subtable}[t]{\linewidth}%
            {%
                    \vspace{-2mm}\begin{tabular}[t]{P{2.5cm}|P{1.2cm}}
                            \toprule
                              Training Data & $\mathcal{J}\&\mathcal{F}$ \\
                            \cmidrule(lr){1-2}
                                            Images  & 57.1\\
                                            Video   & 60.7\\
                                            Images + Video  & 64.4\\
                            \bottomrule
                    \end{tabular}
                \phantomsubcaption{}
                \label{tab:ablation_data}
            }
            \end{subtable}\\
            &\\[-4pt]%
            {\scriptsize\textbf{(d)}}&\begin{subtable}{\linewidth}
            {
              \begin{tabular}[t]{c|c|c}
                           \toprule
                            Object Category & $\mathcal{AP}$ & $\mathcal{AR}$\\
                            \cmidrule(lr){1-3}
                                            Oracle  & 33.0 & 34.5\\
                                            Predicted   & 24.7 & 31.8\\
                            \bottomrule
                    \end{tabular}
                \phantomsubcaption{}
                \label{tab:ablation_semantic}
            }
            \end{subtable}
            \end{tabular}
        }
        \end{subtable}
        \caption{Ablation studies on DAVIS'19 \texttt{val}:  \subref{tab:ablation_embeddings}: Impact of different embedding mixing functions; \subref{tab:ablation_temporal}: Effect of temporal context; \subref{tab:ablation_data}: Analysis of training data; \subref{tab:ablation_semantic}: Impact of Semantic head on a custom validation split of YT-VIS.}
        \label{tab:ablation}
        % \vspace{-2mm}
\end{table}

\PAR{Temporal Window Size:} Next, we investigate the effect of varying the input clip length on DAVIS'19 \texttt{val} ($\phi_\text{xyff}$ is used throughout). 
As shown in \reftab{ablation_temporal}, larger temporal length helps the TSE decoder to predict better quality mask tubes. 
Increasing the input clip length from $T=4$ to $T=16$ improves the $\mathcal{J\&F}$ from 62.2\% to 64.7\%, respectively. Above $T$\,=\,$24$, the performance decreases.

\PAR{Training Data:} For the DAVIS Unsupervised task, we train on image datasets (COCO\,\cite{Lin14ECCV} and Pascal VOC\,\cite{Everingham10IJCV}) in conjunction with video datasets (YT-VIS\,\cite{Yang19ICCV} and DAVIS\,\cite{Caelles19arXiv}). 
As shown in \reftab{ablation_data}, this combination yields $64.4\%$ $\mathcal{J\&F}$ compared to $60.7\%$ when using only video datasets, and $57.1\%$ when using only image datasets. This highlights the benefit of using a combination of image-augmented and video data. For this ablation, $T=8$ and $\phi_\text{xyff}$ were used.

\PAR{Semantic Head:}
Since our network does not produce semantic labels for objects, we adapt it to tasks requiring such labels by adding a semantic segmentation decoder as explained in \refsec{category_classification}.
Here, we compare the quality of our semantic output to an \textit{oracle} by training our network on a custom train-validation split of YT-VIS. This was done because ground truth annotations for the official validation set are not publicly available. The results presented in \reftab{ablation_semantic} show that using \textit{oracle} category labels improves $\mathcal{AP}$ performance by 8.3 from 24.7 to 33.0. This suggests that our results on the official validation set could be further improved by using a better semantic classifier. We leave this for future work.

\subsection{Comparison with state of the art}
\label{sec:sota}

\PARbegin{Video Object Segmentation:} 
\reftab{davis19-bench} summarizes our results on the DAVIS'19 unsupervised validation set. \texttt{OF-Tracker} and \texttt{RI-Tracker} are our own optical flow and re-id baselines which use proposals from a Mask-RCNN\,\cite{He17ICCV} network trained with the same backbone and data as our method (see supplementary for details). Our method ($64.7\%$ $\mathcal{J\&F}$) outperforms these baselines and the other published methods by a significant margin, even though we use a single, proposal-free network.
AGNN\,\cite{Wang19ICCV}, with the second best score of $61.1\%$, uses object proposals from Mask R-CNN\,\cite{He17ICCV} on the salient regions detected by their network, and associates them over time using optical flow. We also list the top entries of the DAVIS'19 Challenge Workshop in \UNPUB{gray}. UnOVOST\,\cite{Zulfikar19CVPRW} achieves a higher score ($67.0\%$), but (i) uses several networks along with heuristic-based post-processing, (ii) is an order of magnitude slower (1 vs. 7FPS), and (iii) is highly tailored to this benchmark. 
To validate this, we adapted UnOVOST to KITTI-MOTS by re-training its networks and optimizing the post-processing parameters with grid search (for further details and analysis of this experiment, we refer to the supplementary). 
As can be seen in \reftab{mots-kitti-bench}, it does not generalize well to the task of Multi-object Tracking and Segmentation. 

% Preview source code for paragraph 0

\begin{table}[t!]
\setlength{\tabcolsep}{4pt} %
\centering
\resizebox{0.95\linewidth}{!}{%
\sisetup{detect-weight=true}
\begin{tabular}{l| c c c|c|ccc|ccc|c }
\toprule
 \multicolumn{12}{c}{DAVIS 2019 Unsupervised} \\
 \midrule
%  Method& OL & PR & OF & RI & $\mathcal{J}$\&$\mathcal{F}$ & $\mathcal{J}$ Mean& $\mathcal{J}$ Recall & $\mathcal{J}$ Decay & $\mathcal{F}$ Mean& $\mathcal{F}$ Recall & $\mathcal{F}$ Decay & fps\\
 Method & P/D & OF & RI & $\mathcal{J}\&\mathcal{F}$ & $\mathcal{J}$ Mean& $\mathcal{J}$ Recall & $\mathcal{J}$ Decay & $\mathcal{F}$ Mean& $\mathcal{F}$ Recall & $\mathcal{F}$ Decay & fps\\
\cmidrule(lr){1-12}
\texttt{\UNPUB{$\text{KIS}^*$~\cite{Cho19CVPRW}}}  & \checkmark &&\checkmark& \UNPUB{59.9} & \UNPUB{-} & \UNPUB{ -  }&\UNPUB{ - }&\UNPUB{ - }&\UNPUB{ - }&\UNPUB{ -} & \UNPUB{-}\\
\texttt{\UNPUB{$\text{UnOVOST}^*$~\cite{Zulfikar19CVPRW}}}  & \checkmark &\checkmark&\checkmark& \UNPUB{67.0} & \UNPUB{65.6} & \UNPUB{ 75.5  }&\UNPUB{ 0.3 }&\UNPUB{ 68.4 }&\UNPUB{ 75.9 }&\UNPUB{3.7} & \UNPUB{$<1$}\\
\cmidrule(lr){1-12}
\texttt{RVOS~\cite{Ventura19CVPR}}  & &&& 41.2 & 36.8 & 40.2 & 0.5 & 45.7 & 46.4 & 1.7 & 20+\\
\texttt{AGNN~\cite{Wang19ICCV}}  & \checkmark&\checkmark&& 61.1 & 58.9&65.7&11.7&63.2&67.1&14.3& -  \\
\cmidrule(lr){1-12}
\texttt{OF-Tracker} & \checkmark&\checkmark&& 54.6&53.4&60.9&-1.3&55.9&63&1.1& $\sim$1\\
\texttt{RI-Tracker} & \checkmark&&\checkmark& 56.9&55.5&63.3&2.7&58.2&64.4&6.4&$<1$\\
% \cmidrule(lr){1-12}
\texttt{\textbf{Ours}}  & &&& \textbf{64.7} & \textbf{61.5} & \textbf{70.4} & \textbf{-4} & \textbf{67.8} & \textbf{75.5} & \textbf{1.2}& 7 \\
\bottomrule
\end{tabular}
}
% \vspace{-1mm}
\caption{DAVIS'19 validation results for the unsupervised track. P/D: Proposals/Detections, OF: optical flow, RI: Re-Id, $^*$ : uses heuristic post-processing.}
% \vspace{-2mm}
\label{tab:davis19-bench}
\end{table}

\PAR{Video Instance Segmentation (VIS):} This task requires object instances to be segmented, tracked and also assigned a category label. We therefore adapt our network to this setting as explained in \refsec{category_classification}.
We train jointly on YT-VIS\,\cite{Yang19ICCV} and augmented images from COCO\,\cite{Lin14ECCV} (for COCO, we only use the $20$ object classes which overlap with the YT-VIS class set).
Since this is a new task with few published works, we compare our method to various baselines and adaptions of existing works from \cite{Yang19ICCV} in \reftab{ytvis-bench}.
As can be seen, our method performs best with respect to all evaluation metrics. Compared to MaskTrack-RCNN~\cite{Yang19ICCV}, we improve the $\mathcal{AP}$ from $30.3$ to $34.6$, even though MaskTrack-RCNN uses a two stage object detector and incorporates additional cues during post-processing. Since MaskTrack-RCNN uses a ResNet-50 backbone, we also applied this backbone to our network and still improve the $\mathcal{AP}$ from $30.3$ to $30.6$.

\begin{table}[t!]
\setlength{\tabcolsep}{6pt}
\centering
% \footnotesize
%\resizebox{\linewidth}{!}{%
\sisetup{detect-weight=true}
\resizebox{0.95\linewidth}{!}{%
\begin{tabular}{l|ccc|c|cccc}
\toprule
 \multicolumn{9}{c}{YouTube Video Instance Segmentation}\\
\midrule
 Method & FF & P/D & OF & $\mathcal{mAP}$ & $\mathcal{AP}@50$ & $\mathcal{AP}@75$& $\mathcal{AR}@1$ & $\mathcal{AR}@10$\\
\cmidrule(lr){1-9}
%\texttt{\UNPUB{Winning Approach\cite{Luiten19ICCVW}}} &\checkmark&\checkmark&\checkmark&\checkmark& 44.8 & 66.4 & 48.9 & 42.7 & 51.7\\
%\cmidrule(lr){1-10}
\texttt{OSMN MaskProp}\,\cite{Yang18CVPR}  & \checkmark&&&23.4&36.5&25.7&28.9&31.1   \\
\texttt{FEELVOS}\,\cite{Voigtlaender19CVPRFeelVOS}  & \checkmark&&&26.9&42.0&29.7&29.9&33.4   \\
\texttt{IoUTracker+}\,\cite{Yang19ICCV}  & &\checkmark&&23.6&39.2&25.5&26.2&30.9   \\
\texttt{OSMN}\,\cite{Yang18CVPR}  & &\checkmark&& 27.5&45.1&29.1&28.6&33.1   \\
\texttt{DeepSORT}\,\cite{Wojke17ICIP}  &&\checkmark&& 26.1&42.9&26.1&27.8&31.3   \\
\texttt{MaskTrack R-CNN}\,\cite{Yang19ICCV}  & & &\checkmark& 30.3 & 51.1 &32.6 & 31.0 & 35.5   \\
\texttt{SeqTracker}\,\cite{Yang19ICCV}  & &&& 27.5&45.7&28.7&29.7&32.5  \\
\cmidrule(lr){1-9}
\texttt{Ours (ResNet-50)}  & &&& 30.6 & 50.7 & 33.5 & 31.6 & 37.1   \\
% \texttt{\textbf{Ours}} &&&& \textbf{35.0} & \textbf{56.0} & \textbf{38.6} & \textbf{34.4} & \textbf{41.7} \\
\texttt{\textbf{Ours}} &&&& \textbf{34.6} & \textbf{55.8} & \textbf{37.9} & \textbf{34.4} & \textbf{41.6} \\
\bottomrule
\end{tabular}
}
%\vspace{-1mm}
\caption{\label{tab:ytvis-bench} YouTube-VIS validation results. P/D: Proposals/Detections, FF: First Frame Proposals, OF: Optical Flow.}
% \vspace{-2mm}
\label{tab:yvis-bench}
\end{table}
%}
\begin{table}[t!]
\setlength{\tabcolsep}{2pt} %
\centering
\resizebox{0.95\linewidth}{!}{%
\sisetup{detect-weight=true}
% \vspace{6mm}

\begin{tabular}{l|ccc|cccc|cccc}
 \multicolumn{12}{c}{\hspace{1cm}}  \\

\toprule
 \multicolumn{12}{c}{KITTI MOTS}  \\
\midrule
&&&&\multicolumn{4}{c|}{Car}&\multicolumn{4}{c}{Pedestrian}\\
Method & P/D & OF & RI & sMOTSA & MOTSA & MOTSP & IDS & sMOTSA & MOTSA & MOTSP & IDS\\
\cmidrule(lr){1-12} 
\UNPUB{\texttt{UnOVOST\cite{Zulfikar19CVPRW}}} & \checkmark & \checkmark & \checkmark  &  50.7 & 60.2 & 85.6 & 151&  33.4  & 47.7& 76.0& 68 \\
%\cmidrule(lr){1-12}
\texttt{MaskRCNN+maskprop\cite{Voigtlaender19CVPR}}& \checkmark & \checkmark &  &  75.1 & 86.6 & 87.1 & - & 45.0 & 63.5 & 75.6& -  \\
\texttt{TrackRCNN\cite{Voigtlaender19CVPR}} & \checkmark &  & \checkmark &  \textbf{76.2}    & \textbf{87.8}   & \textbf{87.2}& 93   & 46.8     & 65.1     & 75.7& 78  \\
%\UNPUB{\texttt{STE\cite{Hu19Arxiv}}} & & \checkmark &  & 61.3    & 46.1   & 80.1   & &-     & -     &-& -  \\ 
\texttt{\textbf{Ours}} & &&& 72.7    & 83.8   & \textbf{87.2}   & \textbf{76}& \textbf{50.4}     & \textbf{66.1}     & \textbf{77.7}& \textbf{14}   \\                
\bottomrule
\end{tabular}
}
%\vspace{-1mm}
\caption{KITTI MOTS validation set results for Car and Pedestrian class. P/D: Proposals/Detections, OF: optical flow, RI: Re-Id.}
\label{tab:mots-kitti-bench}
\end{table}
%\vspace{2mm}

\PAR{Multi-Object Tracking and Segmentation (MOTS):} Here we again adapt the network for category prediction as required by the task formulation. We outline the results of comparing our method with TrackR-CNN\,\cite{Voigtlaender19CVPR} and UnOVOST\,\cite{Zulfikar19CVPRW} in \reftab{mots-kitti-bench}. 
Current top-performing method on MOTS is the two-stage network that extends Mask-RCNN with a re-id head, trained to learn an appearance embedding vector for each object detection, used for data association. 
Our method achieves the highest sMOTSA score ($50.4$) on the \textit{pedestrian} class, but TrackR-CNN performs better on the \textit{car} class. 
However, for both classes, STEm-Seg suffers significantly fewer ID switches (IDS) compared to TrackR-CNN ($76$ vs. $93$ for the \textit{car} class and $14$ vs. $78$ for the \textit{pedestrian} class). 
This measure is of particular interest to the tracking community since it directly reflects temporal association accuracy. 

Similar to UnOVOST\,\cite{Zulfikar19CVPRW}, TrackR-CNN does not generalize well to other tasks. 
To validate this, we retrained TrackR-CNN on the YT-VIS dataset. 
However, the resulting $\mathcal{AP}$ was less than $10$. We assume this is due to TrackR-CNN that is forced to use a $14\times14$ ROI-Align layer\,\cite{He17ICCV} due to memory constraints. This results in coarse segmentation masks which are heavily penalized by the $\mathcal{AP}$ measure. Furthermore, the ReID-based embeddings can only learn an appearance model, which is a limitation in YT-VIS where similar looking objects often occur in the same scene.

\subsection{Segmentation of Salient Regions in Videos}
\label{sec:saliency_seg}
%
%\cite{Song18ECCV,SiamICRA2019,Hou19BMVC,Jain17CVPR,Tokmakov17ICCV,Koh17CVPR,Yang19ICCVAnchorDiff}
\begin{wraptable}{r}{0.476\textwidth}
\setlength{\tabcolsep}{5px}%
\vspace{-19pt}%
\resizebox{1.0\textwidth}{!}{%
\begin{tabular}{l|cc |ccc}%
\toprule
\multicolumn{6}{c}{DAVIS 2016 Unsupervised}\\
\midrule
Method & OF & CRF & $\mathcal{J}$\&$\mathcal{F}$ & $\mathcal{J}$-mean& $\mathcal{F}$-mean \\
\cmidrule(lr){1-6}
\texttt{LVO\cite{Song18ECCV}} &\checkmark &\checkmark& - & 75.9 & 72.1\\
\texttt{PDB\cite{Song18ECCV}} & &\checkmark& - & 77.2 & 74.5\\
\texttt{MotAdapt\cite{SiamICRA2019}}  & && - & 77.2 & 77.4\\
\texttt{3D-CNN\cite{Hou19BMVC}}  &&& - & 78.3 & 77.2\\
% --- Added by Aljosa ---
\texttt{FusionSeg\cite{Jain17CVPR}} & \checkmark & & - & 70.7& 65.3 \\
\texttt{LVO\cite{Tokmakov17ICCV}} & \checkmark & \checkmark & - & 75.9 & 72.1 \\
\texttt{ARP\cite{Koh17CVPR}} & \checkmark &  & - & 76.2& 70.6 \\
\texttt{PDB\cite{Song18ECCV}} &\checkmark &  & - &77.2& 74.5 \\           
\texttt{AD-Net\cite{Yang19ICCVAnchorDiff}}  & && 78.8 & 79.4& 78.2\\
\texttt{AGNN\cite{Wang19ICCV}}  & &\checkmark& 79.9 & \textbf{80.7} & 79.1\\
%\cmidrule(lr){1-6}
\texttt{\textbf{Ours}}  & && \textbf{80.6} & 80.6 &\textbf{80.6}\\
\bottomrule
\end{tabular}
}
% \vspace{-7px}
\caption{
Results on DAVIS'16 val. for the unsupervised track.
OF: optical flow, CRF: post-processing using CRF.
% \vspace{-20px}
}
\label{tab:davis16-bench}
\end{wraptable}
Finally, we apply our method to the DAVIS'16 unsupervised benchmark\,\cite{Perazzi16CVPR}, where the task is to produce a binary segmentation for the salient regions in a given video clip.
Since separating object instances is not required, we simplify our network to one decoder with two output channels trained for binary segmentation using a bootstrapped cross-entropy loss\,\cite{Wu16arXiv} on randomly selected clips from the YT-VIS and DAVIS datasets. 
Although competing methods are specifically engineered for this task,
our simple setup obtains state-of-the-art results (\reftab{davis16-bench}).
We note that while AGNN\,\cite{Wang19ICCV} and AD-Net\,\cite{Yang19ICCVAnchorDiff} use a stronger DeepLabv3~\cite{Chen17ARXIV} backbone, we use additional video data for training. 
This is needed as we work with 3D input volumes. 
Since we do not perform post-processing, we report results from AD-Net~\cite{Yang19ICCVAnchorDiff} without their DAVIS-specific post-processing for a fair comparison.

\section{Conclusion}
We have proposed a novel \textit{bottom-up} approach for instance segmentation in videos which models video clips as 3D space-time volumes and then separates object instances by clustering learned embeddings. We enhanced the feature representation of these embeddings using novel mixing functions which yield considerable performance improvements over existing formulations. We applied our method to multiple, diverse datasets and achieved state-of-the-art results under both category-aware and category-agnostic settings. We further showed that, compared to existing dataset-specific state-of-the-arts, our approach generalizes much better across different datasets. Finally, we validated our network's temporal context learning ability by performing a separate experiment on video saliency detection and showed that our good results also generalize there.

\PAR{Acknowledgements.}
This project was funded, in parts, by ERC Consolidator Grant DeeVise (ERC-2017-COG-773161), EU project CROWDBOT
(H2020-ICT-2017-779942) and the Humboldt Foundation through the Sofja Kovalevskaja
Award. Computing resources for several experiments were granted by RWTH Aachen University under project 'rwth0519'. We thank Sebastian Hennen for help with experiments and Francis Engelmann, Theodora Kontogianni, Paul Voigtlaender, Gulliem Bras{\'o} and Aysim Toker for helpful discussions.

\clearpage
% ---- Bibliography ----
\bibliographystyle{splncs04}
\bibliography{abbrev_short,mybib}

\begin{thebibliography}{100}
\providecommand{\url}[1]{\texttt{#1}}
\providecommand{\urlprefix}{URL }
\providecommand{\doi}[1]{https://doi.org/#1}

\bibitem{Hu19Arxiv}
Anthony~Hu, Alex~Kendall, R.C.: Learning a spatio-temporal embedding for video
  instance segmentation. arxiv preprint arXiv:1912:08969v  (2019)

\bibitem{Van13ICCV}
Van~den Bergh, M., Roig, G., Boix, X., Manen, S., Van~Gool, L.: Online video
  seeds for temporal window objectness. In: ICCV (2013)

\bibitem{Berman18CVPR}
Berman, M., Blaschko, M.B.: Optimization of the jaccard index for image
  segmentation with the lov{\'{a}}sz hinge. CVPR  (2018)

\bibitem{Berman18CVPR2}
Berman, M., Rannen~Triki, A., Blaschko, M.B.: The lov{\'a}sz-softmax loss: a
  tractable surrogate for the optimization of the intersection-over-union
  measure in neural networks. In: CVPR (2018)

\bibitem{Bernardin08JIVP}
Bernardin, K., Stiefelhagen, R.: Evaluating multiple object tracking
  performance: The clear mot metrics. JIVP  \textbf{2008},  1:1--1:10 (2008)

\bibitem{Bochinski17AVSS}
Bochinski, E., Eiselein, V., Sikora, T.: High-speed tracking-by-detection
  without using image information. In: AVSS (2017)

\bibitem{Brox10ECCV}
Brox, T., Malik, J.: Object segmentation by long term analysis of point
  trajectories. In: ECCV (2010)

\bibitem{ButtCollins13CVPR}
Butt, A.A., Collins, R.T.: Multi-target tracking by lagrangian relaxation to
  min-cost network flow. In: CVPR (June 2013)

\bibitem{Caelles17CVPR}
Caelles, S., Maninis, K.K., Pont-Tuset, J., Leal-Taix\'e, L., Cremers, D., {Van
  Gool}, L.: One-shot video object segmentation. In: CVPR (2017)

\bibitem{Caelles18arXiv}
Caelles, S., Montes, A., Maninis, K.K., Chen, Y., {Van Gool}, L., Perazzi, F.,
  Pont-Tuset, J.: The 2018 davis challenge on video object segmentation. arXiv
  preprint arXiv:1803.00557  (2018)

\bibitem{Caelles19arXiv}
Caelles, S., Pont{-}Tuset, J., Perazzi, F., Montes, A., Maninis, K., Gool,
  L.V.: The 2019 {DAVIS} challenge on {VOS:} unsupervised multi-object
  segmentation. arXiv arXiv:1905.00737  (2019)

\bibitem{Chen17ARXIV}
Chen, L., Papandreou, G., Schroff, F., Adam, H.: Rethinking atrous convolution
  for semantic image segmentation. arXiv preprint arXiv:1706.05587  (2017)

\bibitem{Chen19ICCV}
Chen, X., Girshick, R., He, K., Doll{\'a}r, P.: Tensormask: A foundation for
  dense object segmentation. In: ICCV (2019)

\bibitem{Chen2018CVPR}
Chen, Y., Pont-Tuset, J., Montes, A., Van~Gool, L.: Blazingly fast video object
  segmentation with pixel-wise metric learning. In: CVPR (2018)

\bibitem{Cho19CVPRW}
Cho, D., Hong, S., amd J.~Kim, S.K.: Key instance selection for unsupervised
  video object segmentation. The 2019 DAVIS Challenge on Video Object
  Segmentation - CVPR Workshops  (2019)

\bibitem{Comaniciu02TPAMI}
Comaniciu, D., Meer, P.: Mean shift: A robust approach toward feature space
  analysis. PAMI  \textbf{24}(5),  603--619 (2002)

\bibitem{Dave19arxiv}
Dave, A., Tokmakov, P., Ramanan, D.: Towards segmenting everything that moves.
  arXiv preprint arXiv:1902.03715  (2019)

\bibitem{Brabandere17CVPRW}
De~Brabandere, B., Neven, D., Van~Gool, L.: Semantic instance segmentation for
  autonomous driving. In: CVPR Workshops (2017)

\bibitem{Brabandere17arxiv}
De~Brabandere, B., Neven, D., Van~Gool, L.: Semantic instance segmentation with
  a discriminative loss function. arXiv preprint arXiv:1708.02551  (2017)

\bibitem{Dong19ICCVW}
Dong, M., Wang, J., Huang, Y., Yu, D., Su, K., Zhou, K., Shao, J., Wen, S.,
  Wang, C.: Temporal feature augmented network for video instance segmentation.
  In: ICCV Workshops (2019)

\bibitem{Elich19GCPR}
Elich, C., Engelmann, F., Schult, J., Kontogianni, T., Leibe, B.: {3D-BEVIS:
  Birds-Eye-View Instance Segmentation}. In: German Conference on Pattern
  Recognition (GCPR) (2019)

\bibitem{Engelmann20CVPR}
Engelmann, F., Bokeloh, M., Fathi, A., Leibe, B., Nie{\ss}ner, M.: {3D-MPA:
  Multi Proposal Aggregation for 3D Semantic Instance Segmentation}. In: {IEEE
  Conference on Computer Vision and Pattern Recognition (CVPR)} (2020)

\bibitem{Ester96KDD}
Ester, M., Kriegel, H.P., Sander, J., Xu, X., et~al.: A density-based algorithm
  for discovering clusters in large spatial databases with noise. In: ACM
  Conference on Knowledge Discovery and Data Mining (KDD) (1996)

\bibitem{Everingham10IJCV}
Everingham, M., {Van Gool}, L., Williams, C., Winn, J., Zisserman, A.: The
  pascal visual object classes {(VOC)} challenge. IJCV  \textbf{88}(2),
  303--338 (2010)

\bibitem{Feichtenhofer17ICCV}
Feichtenhofer, C., Pinz, A., Zisserman, A.: Detect to track and track to
  detect. In: ICCV (2017)

\bibitem{Feng19ICCVW}
Feng, Q., Yang, Z., Li, P., Wei, Y., Yang, Y.: Dual embedding learning for
  video instance segmentation. In: ICCV Workshops (2019)

\bibitem{Fukunaga1975TOIT}
Fukunaga, K., Hostetler, L.: The estimation of the gradient of a density
  function, with applications in pattern recognition. IEEE Transactions on
  information theory  \textbf{21}(1),  32--40 (1975)

\bibitem{Geiger12CVPR}
Geiger, A., Lenz, P., Urtasun, R.: Are we ready for autonomous driving? the
  {KITTI} vision benchmark suite. In: CVPR (2012)

\bibitem{Gkioxari15CVPR}
Gkioxari, G., Malik, J.: Finding action tubes. In: CVPR (2015)

\bibitem{Gori05IJCNN}
Gori, M., Monfardini, G., Scarselli, F.: A new model for learning in graph
  domains. In: IJCNN (2005)

\bibitem{Han16arxiv}
Han, W., Khorrami, P., Paine, T.L., Ramachandran, P., Babaeizadeh, M., Shi, H.,
  Li, J., Yan, S., Huang, T.S.: Seq-nms for video object detection. arXiv
  preprint arXiv:1602.08465  (2016)

\bibitem{He17ICCV}
He, K., Gkioxari, G., Doll{\'a}r, P., Girshick, R.: Mask {R-CNN}. In: ICCV
  (2017)

\bibitem{He16CVPR}
He, K., Zhang, X., Ren, S., Sun, J.: Deep residual learning for image
  recognition. In: CVPR (2016)

\bibitem{Hermans17ARXIV}
Hermans, A., Beyer, L., Leibe, B.: In defense of the triplet loss for person
  re-identification. arXiv preprint arXiv:1703.07737  (2017)

\bibitem{Hou17ICCV}
Hou, R., Chen, C., Shah, M.: Tube convolutional neural network (t-cnn) for
  action detection in videos. In: ICCV (2017)

\bibitem{Hou19BMVC}
Hou, R., Chen, C., Sukthankar, R., Shah, M.: An efficient 3d {CNN} for
  action/object segmentation in video. In: BMVC (2019)

\bibitem{Ting17NIPS}
Hu, Y., Huang, J., Schwing, A.: Maskrnn: Instance level video object
  segmentation. In: NIPS (2017)

\bibitem{Huang08ECCV}
Huang, C., Wu, B., Nevatia, R.: Robust object tracking by hierarchical
  association of detection responses. In: ECCV (2008)

\bibitem{JainNagel79TPAMI}
Jain, R., Nagel, H.H.: On the analysis of accumulative difference pictures from
  image sequences of real world scenes. PAMI  \textbf{1},  206 -- 214 (1979)

\bibitem{Jain17CVPR}
Jain, S., Xiong, B., Grauman, K.: Fusionseg: Learning to combine motion and
  appearance for fully automatic segmention of generic objects in videos. In:
  CVPR (2017)

\bibitem{jiang20cvpr}
Jiang, L., Zhao, H., Shi, S., Liu, S., Fu, C.W., Jia, J.: Pointgroup: Dual-set
  point grouping for 3d instance segmentation. In: CVPR (2020)

\bibitem{Kang17CVPR}
Kang, K., Li, H., Xiao, T., Ouyang, W., Yan, J., Liu, X., Wang, X.: Object
  detection in videos with tubelet proposal networks. In: CVPR (2017)

\bibitem{Kong18CVPR}
Kong, S., Fowlkes, C.C.: Recurrent pixel embedding for instance grouping. In:
  CVPR (2018)

\bibitem{Kuhn55NRLQ}
Kuhn, H.W., Yaw, B.: The hungarian method for the assignment problem. Naval
  Res. Logist. Quart pp. 83--97 (1955)

\bibitem{Kwak15ICCV}
Kwak, S., Cho, M., Laptev, I., Ponce, J., Schmid, C.: Unsupervised object
  discovery and tracking in video collections. In: ICCV (2015)

\bibitem{LealTaixe15arxiv}
Leal-Taix\'{e}, L., Milan, A., Reid, I., , Roth, S., Schindler, K.:
  {MOTC}hallenge 2015: {T}owards a benchmark for multi-target tracking. arXiv
  preprint arXiv:1504.01942  (2015)

\bibitem{Leibe08IJCV}
Leibe, B., Leonardis, A., Schiele, B.: Robust object detection with interleaved
  categorization and segmentation. IJCV  \textbf{77}(1-3),  259--289 (2008)

\bibitem{Leibe08TPAMI}
Leibe, B., Schindler, K., Cornelis, N., Gool, L.V.: Coupled object detection
  and tracking from static cameras and moving vehicles. PAMI  \textbf{30}(10),
  1683--1698 (2008)

\bibitem{Li18CVPR}
Li, S., Seybold, B., Vorobyov, A., Fathi, A., Huang, Q., Kuo, C.C.J.: Instance
  embedding transfer to unsupervised video object segmentation. In: CVPR (2018)

\bibitem{Lin14ECCV}
Lin, T.Y., Maire, M., Belongie, S., Hays, J., Perona, P., Ramanan, D.,
  Doll{\'a}r, P., Zitnick, C.L.: Microsoft {COCO}: Common objects in context.
  In: ECCV (2014)

\bibitem{Liu2018NIPS}
Liu, R., Lehman, J., Molino, P., Such, F.P., Frank, E., Sergeev, A., Yosinski,
  J.: An intriguing failing of convolutional neural networks and the coordconv
  solution. In: NIPS (2018)

\bibitem{Liu19ICCVW}
Liu, X., Ye, T.: Spatio-temporal attention network for video instance
  segmentation. In: ICCV Workshops (2019)

\bibitem{Lloyd1982TOIT}
Lloyd, S.: Least squares quantization in pcm. IEEE transactions on information
  theory  \textbf{28}(2),  129--137 (1982)

\bibitem{Luiten18ACCV}
Luiten, J., Voigtlaender, P., Leibe, B.: Premvos: Proposal-generation,
  refinement and merging for video object segmentation. In: Asian Conference on
  Computer Vision (2018)

\bibitem{Mcinnes2017OSS}
McInnes, L., Healy, J., Astels, S.: hdbscan: Hierarchical density based
  clustering. Journal of Open Source Software  \textbf{2}(11), ~205 (2017)

\bibitem{Milan16arxiv}
Milan, A., Leal-Taix\'{e}, L., Reid, I., Roth, S., Schindler, K.: {MOT}16: {A}
  benchmark for multi-object tracking. arXiv preprint arXiv:1603.00831  (2016)

\bibitem{Milan15CVPR}
Milan, A., Leal-Taix\'{e}, L., Schindler, K., Reid, I.: Joint tracking and
  segmentation of multiple targets. In: CVPR (2015)

\bibitem{Neuhold17ICCV}
Neuhold, G., Ollmann, T., Bulo, S.R., Kontschieder, P.: The {Mapillary Vistas}
  dataset for semantic understanding of street scenes. In: ICCV (2017)

\bibitem{Neven19CVPR}
Neven, D., Brabandere, B.D., Proesmans, M., Gool, L.V.: Instance segmentation
  by jointly optimizing spatial embeddings and clustering bandwidth. In: CVPR
  (2019)

\bibitem{Newell17NIPS}
Newell, A., Huang, Z., Deng, J.: Associative embedding: End-to-end learning for
  joint detection and grouping. In: NIPS (2017)

\bibitem{Novotny18ECCV}
Novotny, D., Albanie, S., Larlus, D., Vedaldi, A.: Semi-convolutional operators
  for instance segmentation. In: ECCV (2018)

\bibitem{Ochs12CVPR}
Ochs, P., Brox, T.: Higher order motion models and spectral clustering. In:
  CVPR (2012)

\bibitem{Oh2019ICCV}
Oh, S.W., Lee, J.Y., Xu, N., Kim, S.J.: Video object segmentation using
  space-time memory networks. In: ICCV (2019)

\bibitem{Okuma04ECCV}
Okuma, K., Taleghani, A., De~Freitas, N., Little, J.J., Lowe, D.G.: A boosted
  particle filter: Multitarget detection and tracking. In: ECCV (2004)

\bibitem{Osep18ICRA}
O\v{s}ep, A., Mehner, W., Voigtlaender, P., Leibe, B.: Track, then decide:
  Category-agnostic vision-based multi-object tracking. ICRA  (2018)

\bibitem{Osep19ICRA}
O\v{s}ep, A., Voigtlaender, P., Luiten, J., Breuers, S., Leibe, B.: Large-scale
  object mining for object discovery from unlabeled video  (2019)

\bibitem{Osep19arxiv}
O\v{s}ep, A., Voigtlaender, P., Weber, M., Luiten, J., Leibe, B.: 4d generic
  video object proposals. In: ICRA (2020)

\bibitem{Palmer02}
Palmer, S.E.: Organizing objects and scenes. Foundations of cognitive
  psychology: Core readings pp. 189--211 (2002)

\bibitem{Palou13CVPR}
Palou, G., Salembier, P.: Hierarchical video representation with trajectory
  binary partition tree. In: CVPR (2013)

\bibitem{Paragios00TPAMI}
Paragios, N., Deriche, R.: Geodesic active contours and level sets for the
  detection and tracking of moving objects. PAMI  \textbf{22},  266--280 (2000)

\bibitem{Pinheiro16ECCV}
Pinheiro, P., Lin, T., Collobert, R., Doll{\'{a}}r, P.: Learning to refine
  object segments. In: ECCV (2016)

\bibitem{Pinheiro16NIPS}
Pinheiro, P., Collobert, R., Dollár, P.: Learning to segment object
  candidates. In: NIPS (2015)

\bibitem{Perazzi16CVPR}
Pont-Tuset, J., Perazzi, F., Caelles, S., Arbeláez, P., {Sorkine}-{Hornung},
  A., Gool, L.V.: A benchmark dataset and evaluation methodology for video
  object segmentation. In: CVPR (2016)

\bibitem{Qi19CVPR}
Qi, C.R., Litany, O., He, K., Guibas, L.J.: Deep hough voting for 3d object
  detection in point clouds. In: CVPR (2019)

\bibitem{Ren15NIPS}
Ren, S., He, K., Girshick, R., Sun, J.: Faster {R-CNN}: Towards real-time
  object detection with region proposal networks. In: NIPS (2015)

\bibitem{Schroff15CVPR}
Schroff, F., Kalenichenko, D., Philbin, J.: Facenet: {A} unified embedding for
  face recognition and clustering. In: CVPR (2015)

\bibitem{SiamICRA2019}
Siam, M., Jiang, C., Lu, S.W., Petrich, L., Gamal, M., Elhoseiny, M.,
  J{\"{a}}gersand, M.: Video segmentation using teacher-student adaptation in a
  human robot interaction {(HRI)} setting. In: ICRA (2018)

\bibitem{Song18ECCV}
Song, H., Wang, W., Zhao, S., Shen, J., Lam, K.M.: Pyramid dilated deeper
  convlstm for video salient object detection. In: ECCV (September 2018)

\bibitem{Sun18CVPR}
Sun, D., Yang, X., Liu, M.Y., Kautz, J.: {PWC-Net}: {CNNs} for optical flow
  using pyramid, warping, and cost volume. In: CVPR (2018)

\bibitem{Szegedy17AAAI}
Szegedy, C., Ioffe, S., Vanhoucke, V., Alemi, A.A.: Inception-v4,
  inception-resnet and the impact of residual connections on learning. In:
  American Ass. of Art. Intelligence (2017)

\bibitem{Teichman11ICRA}
Teichman, A., Levinson, J., Thrun, S.: Towards {3D} object recognition via
  classification of arbitrary object tracks. In: ICRA (2011)

\bibitem{Tokmakov17ICCV}
Tokmakov, P., Alahari, K., Schmid, C.: Learning video object segmentation with
  visual memory. In: ICCV (2017)

\bibitem{Ventura19CVPR}
Ventura, C., Bellver, M., Girbau, A., Salvador, A., Marqu{\'{e}}s, F.,
  {Gir{'{o}} i Nieto}, X.: {RVOS:} end-to-end recurrent network for video
  object segmentation. CVPR  (2019)

\bibitem{Voigtlaender19CVPRFeelVOS}
Voigtlaender, P., Chai, Y., Schroff, F., Adam, H., Leibe, B., Chen., L.C.:
  Feelvos: Fast end-to-end embedding learning for video object segmentation.
  In: CVPR (2019)

\bibitem{Voigtlaender19CVPR}
Voigtlaender, P., Krause, M., Osep, A., Luiten, J., Sekar, B., Geiger, A.,
  Leibe, B.: {MOTS}: Multi-object tracking and segmentation. In: CVPR (2019)

\bibitem{wang20cvpr}
Wang, H., Luo, R., Maire, M., Shakhnarovich, G.: Pixel consensus voting for
  panoptic segmentation. In: CVPR (2020)

\bibitem{Wang14ECCV}
Wang, L., Hua, G., Sukthankar, R., Xue, J., Zheng, N.: Video object discovery
  and co-segmentation with extremely weak supervision. In: ECCV (2014)

\bibitem{Wang19ICCVW}
Wang, Q., He, Y., Yang, X., Yang, Z., Torr, P.: An empirical study of
  detection-based video instance segmentation. In: In ICCV Workshops (2019)

\bibitem{Wang19ICCV}
Wang, W., Lu, X., Shen, J., Crandall, D.J., Shao, L.: Zero-shot video object
  segmentation via attentive graph neural networks. In: The IEEE International
  Conference on Computer Vision (ICCV) (2019)

\bibitem{Wojke17ICIP}
Wojke, N., Bewley, A., Paulus., D.: Onboard contextual classification of {3D}
  point clouds with learned high-order markov random fields. In: ICIP (2017)

\bibitem{Wren97TPAMI}
Wren, C.R., Azarbayejani, A., Darrell, T., Pentland, A.: Pfinder: Real-time
  tracking of the human body. PAMI  \textbf{19},  780--785 (1997)

\bibitem{Wu18ECCV}
Wu, Y., He, K.: Group normalization. In: ECCV (2018)

\bibitem{Wu16arXiv}
Wu, Z., Shen, C., Hengel, A.v.d.: Wider or deeper: Revisiting the resnet model
  for visual recognition. arXiv preprint arXiv:1611.10080  (2016)

\bibitem{Wug18CVPR}
Wug~Oh, S., Lee, J.Y., Sunkavalli, K., Joo~Kim, S.: Fast video object
  segmentation by reference-guided mask propagation. In: CVPR (2018)

\bibitem{Wug2018CVPR}
Wug~Oh, S., Lee, J.Y., Sunkavalli, K., Joo~Kim, S.: Fast video object
  segmentation by reference-guided mask propagation. In: CVPR (2018)

\bibitem{Xiao16CVPR}
Xiao, F., Jae~Lee, Y.: Track and segment: An iterative unsupervised approach
  for video object proposals. In: CVPR (2016)

\bibitem{xie19cvpr}
Xie, C., Xiang, Y., Harchaoui, Z., Fox, D.: Object discovery in videos as
  foreground motion clustering. In: CVPR (2019)

\bibitem{Xu12CVPR}
Xu, C.: Evaluation of super-voxel methods for early video processing. In: CVPR
  (2012)

\bibitem{Xu18ECCV}
Xu, N., Yang, L., Fan, Y., Yang, J., Yue, D., Liang, Y., Price, B., Cohen, S.,
  Huang, T.: {YouTube-VOS}: Sequence-to-sequence video object segmentation. In:
  ECCV (2018)

\bibitem{Yang19ICCV}
Yang, L., Fan, Y., Xu, N.: Video instance segmentation. In: ICCV (2019)

\bibitem{Yang18CVPR}
Yang, L., Wang, Y., Xiong, X., Yang, J., Katsaggelos, A.K.: Efficient video
  object segmentation via network modulation. In: CVPR (2018)

\bibitem{Yang19ICCVAnchorDiff}
Yang, Z., Wang, Q., Bertinetto, L., Hu, W., Bai, S., Torr, P.H.S.: Anchor
  diffusion for unsupervised video object segmentation. In: ICCV (2019)

\bibitem{Koh17CVPR}
Yeong Jun~Koh, C.S.K.: Primary object segmentation in videos based on region
  augmentation and reduction. In: CVPR (2017)

\bibitem{Yu2015ICML}
Yu, J., Blaschko, M.: Learning submodular losses with the lov{\'a}sz hinge. In:
  International Conference on Machine Learning (ICML) (2015)

\bibitem{Zeng19ICCV}
Zeng, X., Liao, R., Gu, L., Xiong, Y., Fidler, S., Urtasun, R.: Dmm-net:
  Differentiable mask-matching network for video object segmentation. In: ICCV
  (2019)

\bibitem{zhang2020spatial}
Zhang, D., Chun, J., Cha, S.K., Kim, Y.M.: Spatial semantic embedding network:
  Fast 3d instance segmentation with deep metric learning. arXiv preprint
  arXiv:2007.03169  (2020)

\bibitem{Zulfikar19CVPRW}
Zulfikar, I.E., Luiten, J., Leibe, B.: {UnOVOST: Unsupervised Offline Video
  Object Segmentation and Tracking for the 2019 Unsupervised DAVIS Challenge}.
  The 2019 DAVIS Challenge on Video Object Segmentation - CVPR Workshops
  (2019)

\end{thebibliography}

\appendix
\title{Supplementary Material} % Replace with your title
\author{}
\institute{}
\maketitle

%-------------------SECTIONS NUMBERS IN MAIN TEXT-----------------------
\newcommand{\secProblemFormulation}{Sec.~\ref{sec:problem_formulation}\@\xspace}
\newcommand{\secEmbeddingRepresentation}{Sec.~\ref{sec:embedding_representation}\@\xspace}
\newcommand{\secCategoryPrediction}{Sec.~\ref{sec:category_classification}\@\xspace}
\newcommand{\secTraining}{Sec.~\ref{sec:training}\@\xspace}
\newcommand{\secAblations}{Sec.~\ref{sec:ablation}\@\xspace}
\newcommand{\secComparisonToSOTA}{Sec.~\ref{sec:benchmarks}\@\xspace}
\newcommand{\secSaliencyDetection}{Sec.~\ref{sec:saliency_seg}\@\xspace}
%-----------------------------------------------------------------------

% Change section numbering to Roman numbers to differentiate from main text
\renewcommand\thesection{\Roman{section}.}
\renewcommand{\thetable}{\Roman{table}}

\section{Loss Function}
\label{sec:loss}

As explained in \secProblemFormulation of the paper, we use a loss function that is a linear combination of three components:

\begin{equation}
    L_\text{total} = L_\text{emb} + L_\text{smooth} + L_\text{center},
\end{equation}
here $L_\text{smooth}$ is the variance smoothness loss, which ensures that the network outputs uniform variance values for every object instance. For example, if the network outputs the set of variances $\boldsymbol{\mathcal{V}}_j$ for the $j^\text{th}$ instance in a video clip, then the variance smoothness loss for this set of variances is denoted by $L_\text{smooth}^j$ and is computed as:
\begin{equation*}
    L_\text{smooth}^j = \frac{1}{|\boldsymbol{\mathcal{V}}_j |} ~ \sum\limits_{\boldsymbol{v} \in \boldsymbol{\mathcal{V}}_j} \left( \boldsymbol{v} - \boldsymbol{\bar{v}} \right)^2,
\end{equation*}
where $\boldsymbol{\bar{v}}$ is the \textit{mean of the variances} in $\boldsymbol{\mathcal{V}}_j$. Likewise, the loss can be computed for all object instances in the video clip and averaged. No loss is applied to the variances output for background pixels.

$L_\text{center}$ is a regression loss, which ensures that pixels belonging to a foreground object instance have probability values in the instance center heat map $\mathcal{H}$ that match the probability obtained by applying Eq.~2 (main text) to the embedding vector at that pixel location. 

$L_\text{emb}$ is the embedding loss, and is computed using the Lov\`{a}sz extension of the hinge loss for binary segmentation, as explained below.

\PAR{The Lov\`{a}sz Hinge Loss:}
We use the Lov\`{a}sz Hinge Loss~\cite{Berman18CVPR} to train the embeddings output by our network ($L_\text{emb}$). It is a convex surrogate of the Jaccard index which directly optimizes the Intersection over Union (IoU) between the predicted and the ground truth object mask tubes, thereby alleviating class imbalance issues that arise from using, \eg, the cross-entropy loss. In practice, we apply the Lov\`{a}sz Hinge loss for binary segmentation. For a given video object instance prediction, we use $F$ to denote the set of scores for each pixel in the video\footnote{This is practically just the logit value for the probability computed in Eq.~2 (main text).}, and denote by $\Delta_J$ the set of incorrect pixel predictions~\footnote{Obtained by thresholding the probabilities as in Eq.~3 of the main text and comparing against the ground truth mask.}. The loss $L_\text{emb}$ can then be computed as follows:
\begin{equation}
    L_\text{emb}(F) = \bar{\Delta_J}(h(F)),
\end{equation}
where $\bar{\Delta_J}$ is the Lov\`{a}sz extension of ${\Delta_J}$, and $h$ is the hinge loss associated with a binary prediction.
Here we provide only a high-level description of the loss function. For a more detailed explanation of this loss we refer the reader to \cite{Berman18CVPR}.
%
% \TODO{aljosa: honestly, the explanation is not very clear}

\section{Implementation Details}
\label{sec:training_setup}

\PAR{Hardware:} We train our network using a batch size of 2 on a workstation with 2 Nvidia RTX TITAN GPUs and 64GB RAM. All inference experiments were performed on a workstation with a single Nvidia GTX 1080Ti GPU and 32GB RAM. 

\PAR{Training Schedule:} For all tasks, the network is trained using an SGD optimizer with an initial learning rate of $10^{-3}$. The learning rate is initially constant and then starts to decay exponentially after a certain number of iterations up to $10^{-5}$. The exact number of iterations varies for each setting as follows:

\begin{itemize}
    \item DAVIS'19 Unsupervised: 60k total iterations, decay begins after 20k iterations.
    \item YouTube-VIS: 150k total iterations, decay begins after 60k iterations.
    \item KITTI-MOTS: 100k total iterations, decay begins after 40k iterations.
\end{itemize}

\PAR{Image Augmented Sequences:} As mentioned in~\secTraining, we train our network on clips from actual video data in addition to sequences that have been synthesized from static images using random affine transformations and motion blur. Doing so allows us to utilize a large amount of publicly available image instance segmentation data (\eg, COCO~\cite{Lin14ECCV}, Pascal-VOC~\cite{Everingham10IJCV}, Mapillary Vistas~\cite{Neuhold17ICCV}) for training purposes. We experimentally verified the performance benefit of incorporating such data in~\secAblations. 

These augmentations were applied using the  \texttt{imgaug}~\footnote{https://github.com/aleju/imgaug} library, which, in addition to various transformations, also provides a built-in function that simulates image blur caused by camera motion. The affine transformations we apply consist of rotations in the range $[\ang{-10}, \ang{10}]$, translations of up to $10\%$ of the image dimension along each axis, and scale variations in the range $[0.8, 1.2]$. We also apply small random offsets to the hue and saturation values of each image. All random transformations are independent of one another, \ie, we do not try to simulate consistent motion by sequentially applying the same transformation multiple times.

\PAR{Video Data Augmentation:} For training clips sampled from actual video data, random horizontal flipping is the only augmentation used. This is applied randomly to entire clips and not to individual frames within clips. 

\section{Baselines for DAVIS'19 Unsupervised}
\label{sec:davis19_baselines}

In~\secComparisonToSOTA we compared our method to two simple proposal-based baselines: optical flow tracker (\texttt{OF-Tracker}) and Re-ID tracker (\texttt{RI-Tracker}), in addition to other published methods on DAVIS'19 unsupervised benchmark. 
For both, we generate per-frame mask proposals $M \in \{m_1,...,m_n\}$ for all the objects in a video using a ResNet-101 based Mask R-CNN~\cite{He17ICCV}. 
To ensure a fair comparison with our approach, we train the Mask R-CNN jointly on YouTube-VIS~\cite{Yang19ICCV}, DAVIS'19~\cite{Caelles19arXiv}, and augmented images from COCO~\cite{Lin14ECCV}, as well as Pascal-VOC~\cite{Everingham10IJCV} dataset for 120k iterations. This network is initialized with weights from a model trained for image instance segmentation on COCO. We use SGD with a momentum of 0.9 and an initial learning rate of $10^{-3}$ with exponential decay.
The mask proposals generated by this re-trained Mask R-CNN network are then linked over time using optical flow and re-id for \texttt{OF-Tracker} and \texttt{RI-Tracker}, respectively.

\PAR{\texttt{OF-Tracker}:}
We use PWC-Net~\cite{Sun18CVPR} to generate optical flow for each subsequent pair of frames in the DAVIS'19 validation set. The optical flow is then used to warp $m_{i+1}$ onto $m_i$ for each frame pair $\{i, i+1\}$ to generate a set of warped masks per-frame $W \in \{w_1,...,w_{n-1}\}$ for a video sequence. A simple linear assignment based on object overlap between the warped frame $w_i$ and the proposal $m_i$ is then used to associate the objects in the adjacent video frames. 
The associated object IDs are further propagated forward throughout the video sequence. 

\PAR{\texttt{RI-Tracker}:}
For the \texttt{RI-Tracker}, we train a re-id network with a ResNet-50~\cite{He16CVPR} backbone on the DAVIS'19 \cite{Caelles19arXiv} training set. The network is trained using a batch hard triplet loss~\cite{Hermans17ARXIV} on randomly selected triplets from a random video sequence for 25k iterations. This network is then used to generate re-id vectors for all the object proposals in $M$, which are further associated over time using linear assignment based on the Euclidean distance between embedding vectors.

\section{Extended Ablations for Embedding Mixing Function} %  for YouTube-VIS and KITTI-MOTS
\label{extended_results_ytvis_mots}

In~\secAblations, we ablated the impact of using different mixing functions $\phi(\cdot)$ that modify the embedding representation as discussed in ~\secEmbeddingRepresentation. In Tab.~1(a) of the main text, we reported the results of this ablation on the DAVIS'19 Unsupervised validation set. Here, we provide extended results of applying different $\phi(\cdot)$ on the YouTube-VIS~\cite{Yang19ICCV} and KITTI-MOTS~\cite{Voigtlaender19CVPR} datasets in Tab.~\ref{tab:ablation_supplementary}. The results for the DAVIS'19 Unsupervised validation set have also been repeated for reference.
\begin{table}[t]
    \centering
    \footnotesize
    \vspace{-2mm}
                \begin{tabular}[t]{p{2.5cm} | c | c | c | c | c}
\toprule
            \multirow{2}{2.5cm}{{\footnotesize Mixing Function}} & \multirow{2}{0.5cm}{~~$E$~~} & ~~DAVIS~~  & ~~YT-VIS~~ & \multicolumn{2}{c}{KITTI MOTS} \\
            \cmidrule(lr){3-6}
              && $\mathcal{J}$\&$\mathcal{F}$ & $\mathcal{AP}$ & sMOTSA (\textit{car}) & sMOTSA (\textit{pedestrian}) \\
            \cmidrule(lr){1-6}
                            $\phi_{\text{xy}}$   & 2  & 61.6 & 30.5 & 64.2 & 41.1 \\
                            $\phi_\text{xyt}$  & 3   & 62.6  & 31.8 & 72.5 & \textbf{48.9} \\
                            $\phi_\text{xyf}$  & 3   & 62.8  & 32.6 & 71.8 & 42.2 \\
                            $\phi_\text{xytf}$  & 4  & 64.2  & 32.4 & 71.9 & 43.6 \\
                            $\phi_\text{xyff}$  & 4   & \textbf{64.4} & \textbf{34.6} & 73.2 & 47.3 \\
                            $\phi_\text{xyfff}$ & 5 & 62.4   & 34.0 & \textbf{73.4} & 41.5 \\
            \bottomrule
            \end{tabular}
            \label{tab:ablation_embeddings_supplementary}
        \caption{Ablation studies on the Impact of different embedding mixing functions on DAVIS '19, YouTube-VIS (YT-VIS) and KITTI MOTS.}
        \label{tab:ablation_supplementary}
\end{table}

It can be seen that both DAVIS'19 and YouTube-VIS give consistent results: for the same number of total embedding dimensions ($E$), having a free dimension is more beneficial than having a temporal coordinate dimension. For KITTI-MOTS, however, the trend differs. In particular, we obtain similar performance with $\phi_\text{xyt}$ ($72.5$ and $48.9$ sMOTSA on the \textit{car} and \textit{pedestrian} class, respectively) and $\phi_\text{xyff}$ ($73.2$ and $47.3$ sMOTSA). In Tab.~4 (main text), we reported the results for $\phi_\text{xyt}$ since the mean sMOTSA score for the two categories ($60.70$) is slightly better than that of $\phi_\text{xyff}$ ($60.25$). We attribute this difference in part to the fact that the temporal coordinate is a more useful feature for instance separation in KITTI-MOTS than in DAVIS'19 due to the fact that object instances undergo faster motion and often enter/exit the scene mid-way through a video clip. Furthermore, the performance trends for the \textit{car} and \textit{pedestrian} classes seem to follow different patterns, \eg, while $\phi_\text{xyfff}$ yields the highest sMOTSA for the \textit{car} class ($73.4$), it is significantly lower for the \textit{pedestrian} class. 

\section{UnOVOST Training on KITTI-MOTS}

In~\secComparisonToSOTA, we reported the performance of UnOVOST~\cite{Zulfikar19CVPRW}, the highest-scoring workshop submission for the DAVIS'19 Unsupervised Challenge~\cite{Caelles19arXiv}, for the task of Multi-object Tracking and Segmentation (MOTS) using the KITTI-MOTS dataset~\cite{Voigtlaender19CVPR}.
We obtained the implementation from the authors~\cite{Zulfikar19CVPRW} and re-trained and tuned the model as follows:

\begin{itemize}
    \item We initialized a Mask R-CNN~\cite{He17ICCV} network with a ResNet-101~\cite{Szegedy17AAAI} backbone with weights from an off-the-shelf model trained for instance segmentation on the COCO dataset~\cite{Lin14ECCV}. We then altered the output layers to predict two categories, \ie. \textit{car} and \textit{pedestrian}, and trained the network for 60k iterations on Mapillary Vistas~\cite{Neuhold17ICCV} and KITTI-MOTS datasets. 
    The training data and the backbone is thus identical to the one used for our STEm-Seg network.
    
    \item We trained a ReID network on image instance crops from KITTI-MOTS using a triplet loss~\cite{Schroff15CVPR} and \textit{batch-hard sampling}~\cite{Hermans17ARXIV}.
\end{itemize}
The two most important hyper-parameters in UnOVOST are the IoU thresholds used for pruning object detections and for associating object detections based on optical flow, respectively. We performed a grid search for these two parameters on the KITTI-MOTS validation set in order to optimize the sMOTSA score. Our observation was that the UnOVOST framework is fairly insensitive to these parameters; however, the final scores on KITTI-MOTS are consistently low (see Tab. 4 in the paper). 
Qualitative analysis of the results showed that the ReID network frequently makes spurious associations. 
We postulate that this is because object instances in KITTI-MOTS frequently have similar appearances. This differs from the object instances in DAVIS whose appearances usually differ since they span a large variety of object classes. 

%Since the framework involves multiple components (object detection network, optical flow network, ReID network, optimization for tracklet association), further analysis would require comprehensive ablations for each of these components.

\section{Adaptation of TrackR-CNN to YouTube-VIS}
\label{sec:ytvis}

As discussed in \secComparisonToSOTA, we adapted the publicly available implementation~\footnote{https://github.com/VisualComputingInstitute/TrackR-CNN} of TrackR-CNN~\cite{Voigtlaender19CVPR} to the task of Video Instance Segmentation and evaluated it on the Youtube-VIS dataset~\cite{Yang19ICCV}. 
To this end, we initialized the parameters of the network, which overlap with Mask R-CNN~\cite{He17ICCV} with weights from a model trained for instance segmentation on COCO~\cite{Lin14ECCV} and Mapillary Vistas~\cite{Neuhold17ICCV}.

In the original implementation, a class-specific re-identification embedding head was used. This was feasible for KITTI-MOTS, where there are only two object classes. 
%and lengthy video sequences which each contain sufficient instances belonging to both classes. 
In YouTube-VIS, however, there are 40 object classes, and several occur infrequently in the dataset. Furthermore, video sequences are significantly shorter, and there are usually only 1--2 objects of the same class present in a video clip. For that reason, we adapted the TrackR-CNN architecture and kept a single ReID head that is shared among all object classes. We trained the network under this setting using a batch size of 8 images for 400k iterations and evaluated multiple intermediate checkpoints. Despite these efforts, the highest $\mathcal{AP}$ score obtained was less than $10\%$. 

A major performance bottleneck we identified is a low-resolution 14x14 RoI-Align~\cite{He17ICCV} layer used in TrackR-CNN that limit the memory usage to a reasonable level. This suffices for KITTI-MOTS, which contains small pedestrian instances and cars with simple shapes, but results in very coarse segmentation masks on the YouTube-VIS dataset which contains a diverse set of objects that cover a large area of the image. The $\mathcal{AP}$ measure heavily penalizes such coarse segmentation as it is computed by taking the average over a set of IoU thresholds ranging from 0.5 to 0.95.

% \section{Workshop Submissions for YouTube-VIS}
% OPTIONAL: Should we discuss the winning entry?
% ALJOSA: I would scratch this whole sec, we do not discuss this in the paper.

\section{Additional Qualitative Results}
\label{sec:qualitative}

\begin{figure}[t]
\centering
  \includegraphics[width=0.23\textwidth, frame]{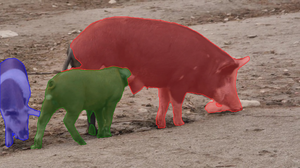}\hspace{1px}%
  \includegraphics[width=0.23\textwidth, frame]{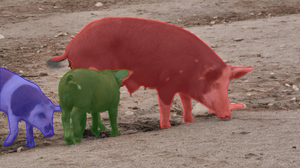}\hspace{1px}%
  \includegraphics[width=0.23\textwidth, frame]{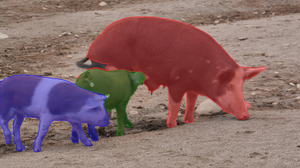}\hspace{1px}%
  \includegraphics[width=0.23\textwidth, frame]{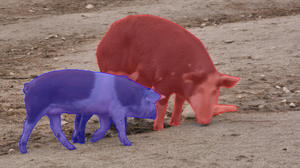}\\
  \includegraphics[width=0.23\textwidth, frame]{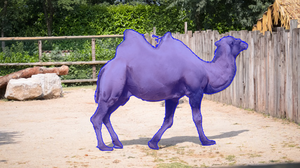}\hspace{1px}%
  \includegraphics[width=0.23\textwidth, frame]{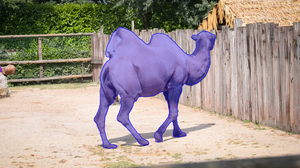}\hspace{1px}%
  \includegraphics[width=0.23\textwidth, frame]{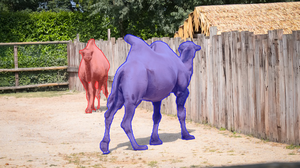}\hspace{1px}%
  \includegraphics[width=0.23\textwidth, frame]{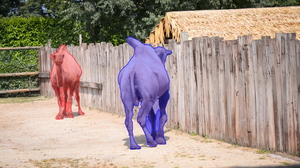}\\
  \includegraphics[width=0.23\textwidth, frame]{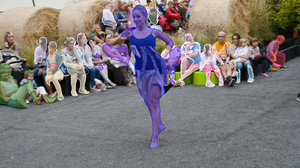}\hspace{1px}%
  \includegraphics[width=0.23\textwidth, frame]{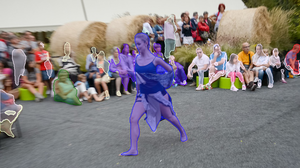}\hspace{1px}%
  \includegraphics[width=0.23\textwidth, frame]{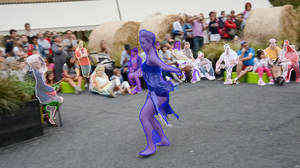}\hspace{1px}%
  \includegraphics[width=0.23\textwidth, frame]{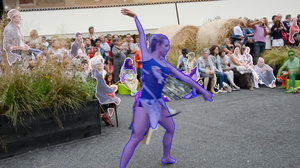}\\
  \includegraphics[width=0.23\textwidth, frame]{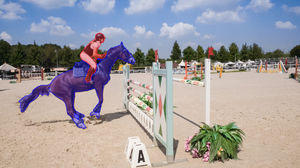}\hspace{1px}%
  \includegraphics[width=0.23\textwidth, frame]{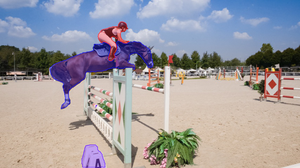}\hspace{1px}%
  \includegraphics[width=0.23\textwidth, frame]{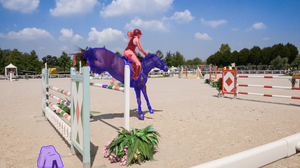}\hspace{1px}%
  \includegraphics[width=0.23\textwidth, frame]{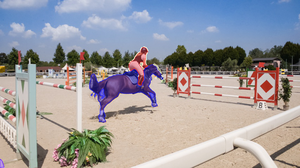}\\
  \includegraphics[width=0.23\textwidth, frame]{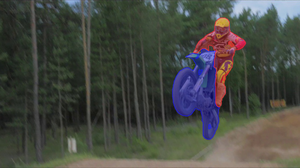}\hspace{1px}%
  \includegraphics[width=0.23\textwidth, frame]{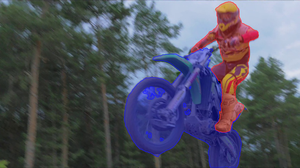}\hspace{1px}%
  \includegraphics[width=0.23\textwidth, frame]{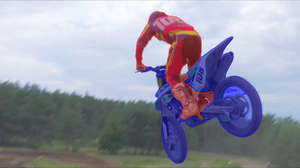}\hspace{1px}%
  \includegraphics[width=0.23\textwidth, frame]{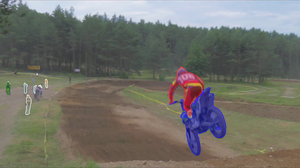}\\
  \includegraphics[width=0.23\textwidth, frame]{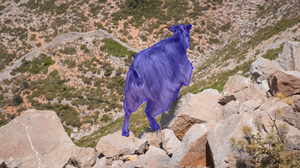}\hspace{1px}%
  \includegraphics[width=0.23\textwidth, frame]{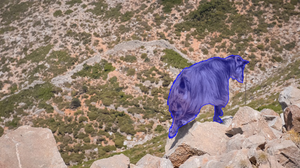}\hspace{1px}%
  \includegraphics[width=0.23\textwidth, frame]{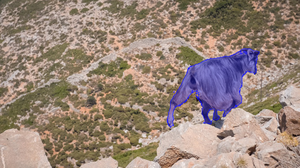}\hspace{1px}%
  \includegraphics[width=0.23\textwidth, frame]{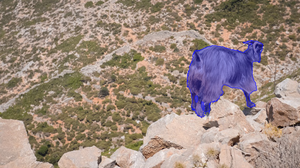}\\
  \includegraphics[width=0.23\textwidth, frame]{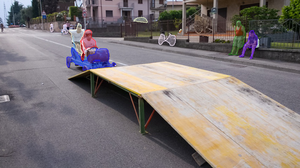}\hspace{1px}%
  \includegraphics[width=0.23\textwidth, frame]{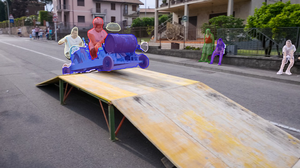}\hspace{1px}%
  \includegraphics[width=0.23\textwidth, frame]{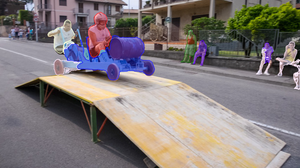}\hspace{1px}%
  \includegraphics[width=0.23\textwidth, frame]{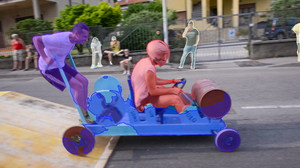}\\
  \includegraphics[width=0.23\textwidth, frame]{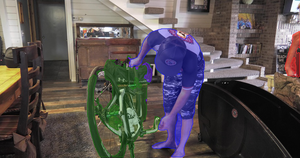}\hspace{1px}%
  \includegraphics[width=0.23\textwidth, frame]{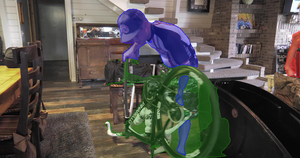}\hspace{1px}%
  \includegraphics[width=0.23\textwidth, frame]{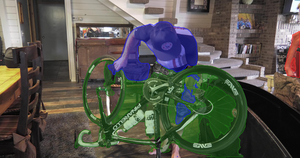}\hspace{1px}%
  \includegraphics[width=0.23\textwidth, frame]{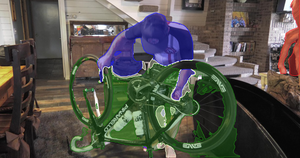}\\
\raggedleft
\begin{tikzpicture}[node distance=2cm]
\node (A) at (2.75, 0) {};
\node (B) at (13.0, 0) {};
\draw[-{Stealth}, to path={-- (\tikztotarget)}](A) edge (B);
\node[text width=1cm] at (13.5,0){time};
\end{tikzpicture}
\vspace{-2px}
  \caption{\textbf{Additional qualitative results on DAVIS'19}. STEm-Seg generates consistently good results under varied scenarios. E.g., in the \texttt{motocross-jump} sequence (\textit{fifth row}) it demonstrates robustness to a large change in scale. In the \texttt{bike-packing} sequence (\textit{bottom row})}, it is robust to sudden pose changes.
    \label{fig:qualitative_davis}
\end{figure}

\begin{figure}[t]
\centering
  \includegraphics[width=0.23\textwidth, trim={0 0 0 0}, clip, frame]{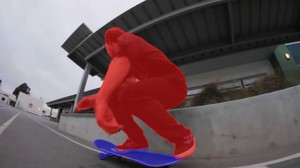}\hspace{1px}%
  \includegraphics[width=0.23\textwidth, trim={0 0 0 0}, clip,frame]{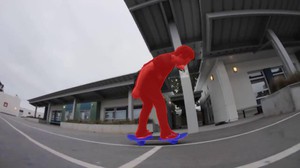}\hspace{1px}%
  \includegraphics[width=0.23\textwidth, trim={0 0 0 0}, clip,frame]{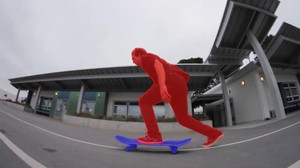}\hspace{1px}%
  \includegraphics[width=0.23\textwidth, trim={0 0 0 0}, clip,frame]{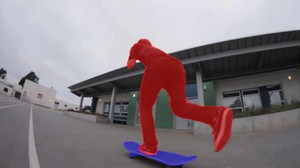}\\
  \includegraphics[width=0.23\textwidth, trim={0 0 0 0}, clip, frame]{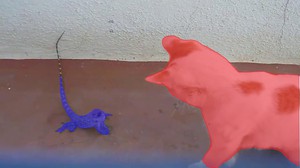}\hspace{1px}%
  \includegraphics[width=0.23\textwidth, trim={0 0 0 0}, clip,frame]{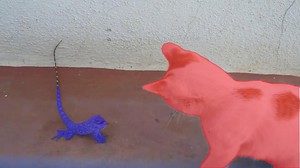}\hspace{1px}%
  \includegraphics[width=0.23\textwidth, trim={0 0 0 0}, clip,frame]{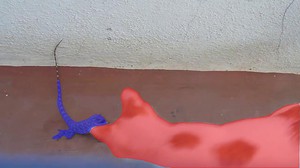}\hspace{1px}%
  \includegraphics[width=0.23\textwidth, trim={0 0 0 0}, clip,frame]{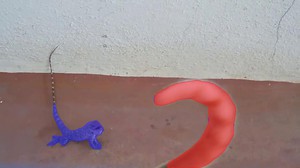}\\
  \includegraphics[width=0.23\textwidth, trim={100 100 0 0}, clip, frame]{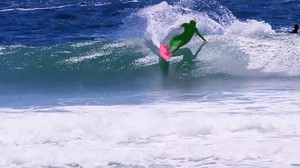}\hspace{1px}%
  \includegraphics[width=0.23\textwidth, trim={100 100 0 0}, clip,frame]{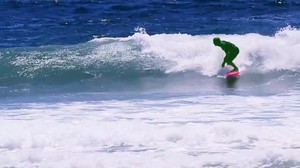}\hspace{1px}%
  \includegraphics[width=0.23\textwidth, trim={100 100 0 0}, clip,frame]{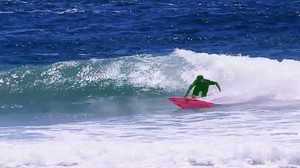}\hspace{1px}%
  \includegraphics[width=0.23\textwidth, trim={100 100 0 0}, clip,frame]{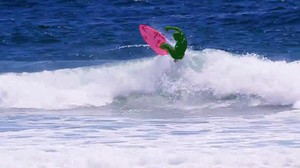}\\
  \includegraphics[width=0.23\textwidth, trim={0 0 0 0}, clip, frame]{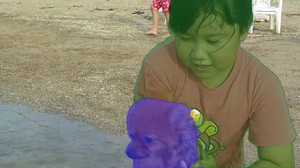}\hspace{1px}%
  \includegraphics[width=0.23\textwidth, trim={0 0 0 0}, clip,frame]{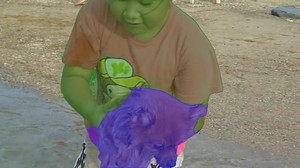}\hspace{1px}%
  \includegraphics[width=0.23\textwidth, trim={0 0 0 0}, clip,frame]{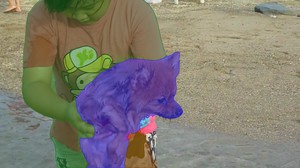}\hspace{1px}%
  \includegraphics[width=0.23\textwidth, trim={0 0 0 0}, clip,frame]{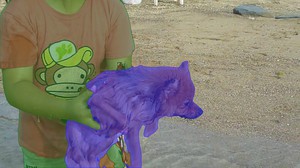}\\
  \includegraphics[width=0.23\textwidth, trim={0 0 0 0}, clip, frame]{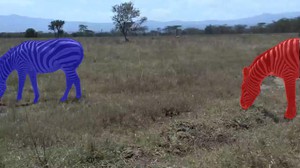}\hspace{1px}%
  \includegraphics[width=0.23\textwidth, trim={0 0 0 0}, clip,frame]{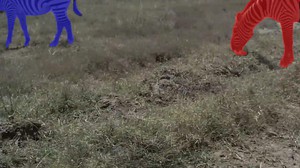}\hspace{1px}%
  \includegraphics[width=0.23\textwidth, trim={0 0 0 0}, clip,frame]{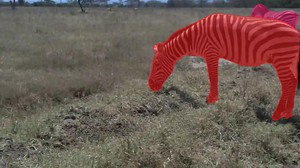}\hspace{1px}%
  \includegraphics[width=0.23\textwidth, trim={0 0 0 0}, clip,frame]{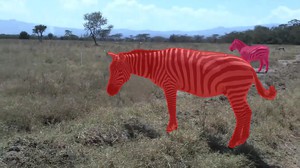}\\
  \includegraphics[width=0.23\textwidth, trim={0 0 0 0}, clip, frame]{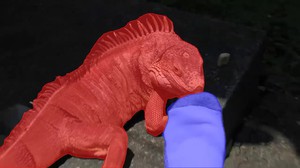}\hspace{1px}%
  \includegraphics[width=0.23\textwidth, trim={0 0 0 0}, clip,frame]{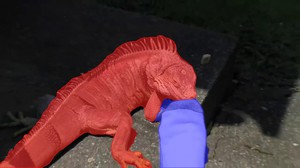}\hspace{1px}%
  \includegraphics[width=0.23\textwidth, trim={0 0 0 0}, clip,frame]{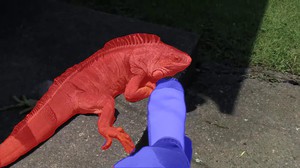}\hspace{1px}%
  \includegraphics[width=0.23\textwidth, trim={0 0 0 0}, clip,frame]{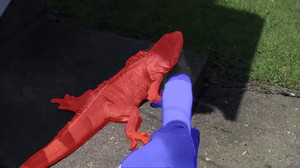}\\
  \includegraphics[width=0.23\textwidth, trim={0 0 0 0}, clip, frame]{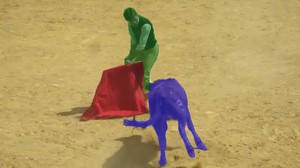}\hspace{1px}%
  \includegraphics[width=0.23\textwidth, trim={0 0 0 0}, clip,frame]{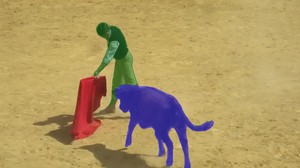}\hspace{1px}%
  \includegraphics[width=0.23\textwidth, trim={0 0 0 0}, clip,frame]{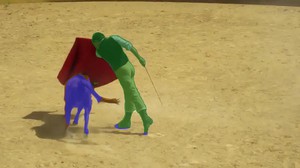}\hspace{1px}%
  \includegraphics[width=0.23\textwidth, trim={0 0 0 0}, clip,frame]{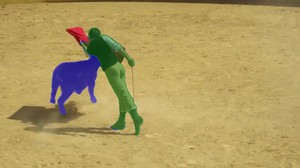}\\
  \includegraphics[width=0.23\textwidth, trim={0 0 0 0}, clip, frame]{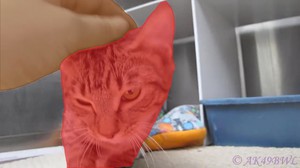}\hspace{1px}%
  \includegraphics[width=0.23\textwidth, trim={0 0 0 0}, clip,frame]{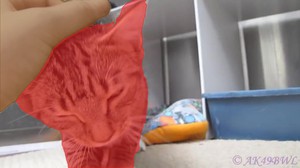}\hspace{1px}%
  \includegraphics[width=0.23\textwidth, trim={0 0 0 0}, clip,frame]{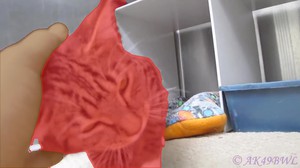}\hspace{1px}%
  \includegraphics[width=0.23\textwidth, trim={0 0 0 0}, clip,frame]{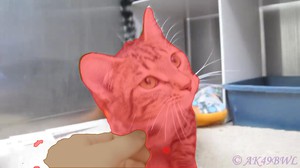}\\
  \includegraphics[width=0.23\textwidth, trim={0 0 0 0}, clip, frame]{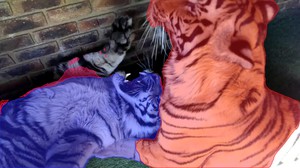}\hspace{1px}%
  \includegraphics[width=0.23\textwidth, trim={0 0 0 0}, clip,frame]{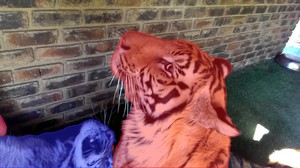}\hspace{1px}%
  \includegraphics[width=0.23\textwidth, trim={0 0 0 0}, clip,frame]{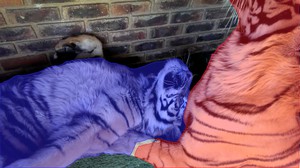}\hspace{1px}%
  \includegraphics[width=0.23\textwidth, trim={0 0 0 0}, clip,frame]{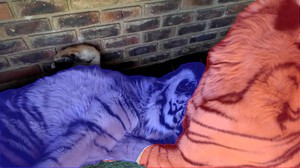}\\
  \includegraphics[width=0.23\textwidth, trim={0 0 0 0}, clip, frame]{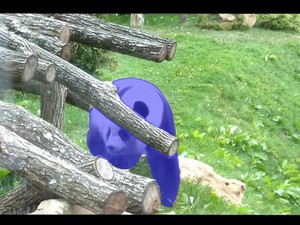}\hspace{1px}%
  \includegraphics[width=0.23\textwidth, trim={0 0 0 0}, clip,frame]{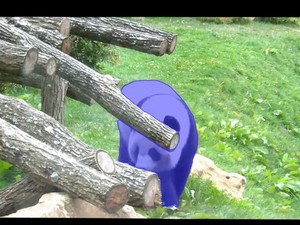}\hspace{1px}%
  \includegraphics[width=0.23\textwidth, trim={0 0 0 0}, clip,frame]{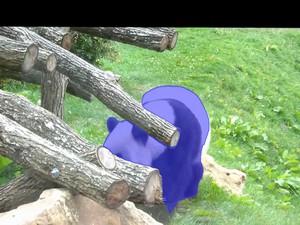}\hspace{1px}%
  \includegraphics[width=0.23\textwidth, trim={0 0 0 0}, clip,frame]{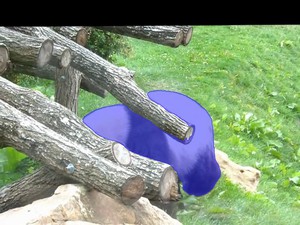}\\
\raggedleft
\begin{tikzpicture}[node distance=2cm]
\node (A) at (2.75, 0) {};
\node (B) at (13.0, 0) {};
\draw[-{Stealth}, to path={-- (\tikztotarget)}](A) edge (B);
\node[text width=1cm] at (13.5,0){time};
\end{tikzpicture}
\vspace{-2px}
  \caption{\textbf{Additional qualitative results on YouTube-VIS (YT-VIS)}~\cite{Yang19ICCV}. Most of the semantically challenging animal categories are successfully segmented by STEm-Seg. It also captures some fine object details such as the skateboard (\textit{top row}) and the surfboard (\textit{third row}) well. }
   \label{fig:qualitative_ytvis}
\end{figure}

\begin{figure}[t]
\centering

  \includegraphics[width=0.23\textwidth,frame]{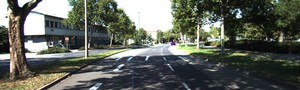}\hspace{1px}%
  \includegraphics[width=0.23\textwidth,frame]{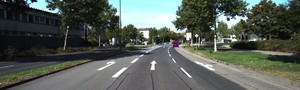}\hspace{1px}%
  \includegraphics[width=0.23\textwidth,frame]{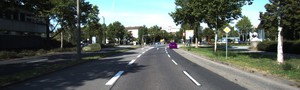}\hspace{1px}%
  \includegraphics[width=0.23\textwidth,frame]{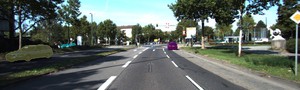}\\ %
  \includegraphics[width=0.23\textwidth,frame]{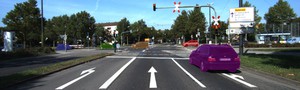}\hspace{1px}%
  \includegraphics[width=0.23\textwidth,frame]{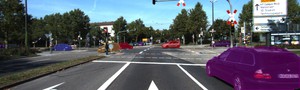}\hspace{1px}%
  \includegraphics[width=0.23\textwidth,frame]{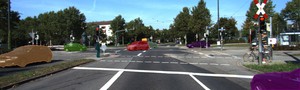}\hspace{1px}%
  \includegraphics[width=0.23\textwidth,frame]{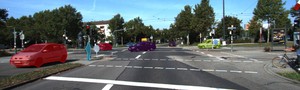}\\ %
  \includegraphics[width=0.23\textwidth,frame]{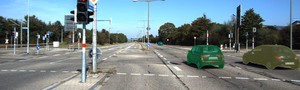}\hspace{1px}%
  \includegraphics[width=0.23\textwidth,frame]{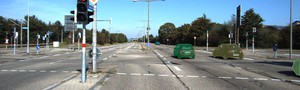}\hspace{1px}%
  \includegraphics[width=0.23\textwidth,frame]{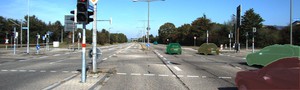}\hspace{1px}%
  \includegraphics[width=0.23\textwidth,frame]{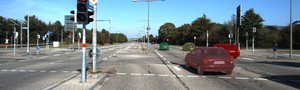}\\
  \includegraphics[width=0.23\textwidth,frame]{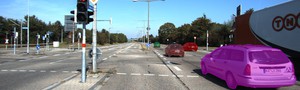}\hspace{1px}%
  \includegraphics[width=0.23\textwidth,frame]{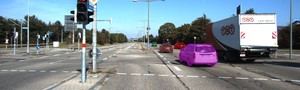}\hspace{1px}%
  \includegraphics[width=0.23\textwidth,frame]{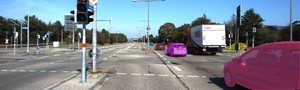}\hspace{1px}%
  \includegraphics[width=0.23\textwidth,frame]{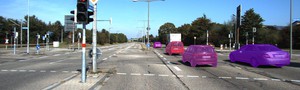}\\
  \includegraphics[width=0.23\textwidth,frame]{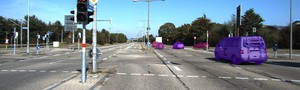}\hspace{1px}%
  \includegraphics[width=0.23\textwidth,frame]{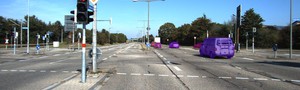}\hspace{1px}%
  \includegraphics[width=0.23\textwidth,frame]{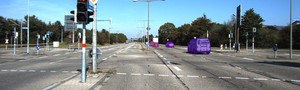}\hspace{1px}%
  \includegraphics[width=0.23\textwidth,frame]{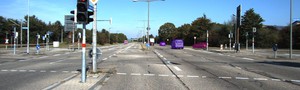}\\
  \includegraphics[width=0.23\textwidth,frame]{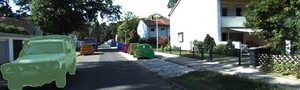}\hspace{1px}%
  \includegraphics[width=0.23\textwidth,frame]{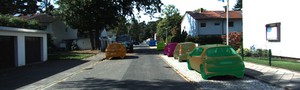}\hspace{1px}%
  \includegraphics[width=0.23\textwidth,frame]{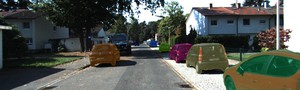}\hspace{1px}%
  \includegraphics[width=0.23\textwidth,frame]{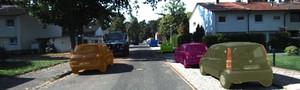}\\ %
  \includegraphics[width=0.23\textwidth,frame]{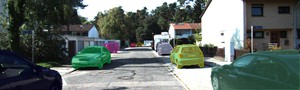}\hspace{1px}%
  \includegraphics[width=0.23\textwidth,frame]{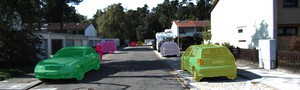}\hspace{1px}%
  \includegraphics[width=0.23\textwidth,frame]{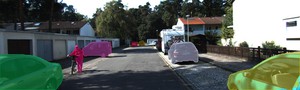}\hspace{1px}%
  \includegraphics[width=0.23\textwidth,frame]{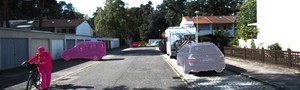}\\%
  \includegraphics[width=0.23\textwidth,frame]{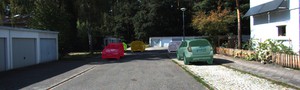}\hspace{1px}%
  \includegraphics[width=0.23\textwidth,frame]{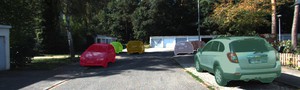}\hspace{1px}%
  \includegraphics[width=0.23\textwidth,frame]{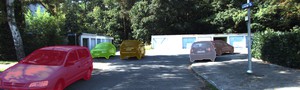}\hspace{1px}%
  \includegraphics[width=0.23\textwidth,frame]{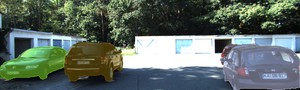}\\%
  \includegraphics[width=0.23\textwidth,frame]{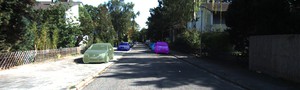}\hspace{1px}%
  \includegraphics[width=0.23\textwidth,frame]{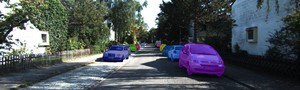}\hspace{1px}%
  \includegraphics[width=0.23\textwidth,frame]{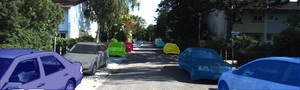}\hspace{1px}%
  \includegraphics[width=0.23\textwidth,frame]{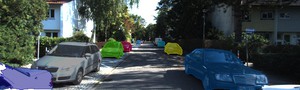}\\%
  \includegraphics[width=0.23\textwidth,frame]{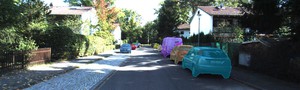}\hspace{1px}%
  \includegraphics[width=0.23\textwidth,frame]{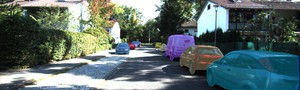}\hspace{1px}%
\includegraphics[width=0.23\textwidth,frame]{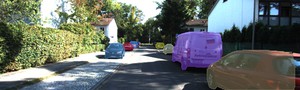}\hspace{1px}%
  \includegraphics[width=0.23\textwidth,frame]{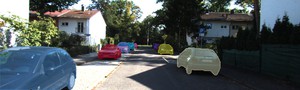}\\%
  \includegraphics[width=0.23\textwidth,frame]{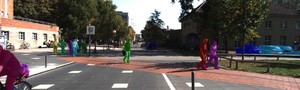}\hspace{1px}%
  \includegraphics[width=0.23\textwidth,frame]{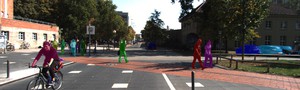}\hspace{1px}%
  \includegraphics[width=0.23\textwidth,frame]{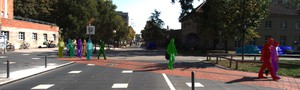}\hspace{1px}%
  \includegraphics[width=0.23\textwidth,frame]{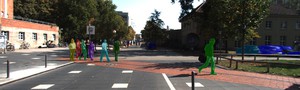}\\%
  \includegraphics[width=0.23\textwidth,frame]{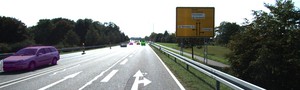}\hspace{1px}%
  \includegraphics[width=0.23\textwidth,frame]{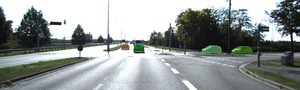}\hspace{1px}%
  \includegraphics[width=0.23\textwidth,frame]{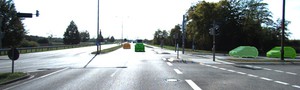}\hspace{1px}%
  \includegraphics[width=0.23\textwidth,frame]{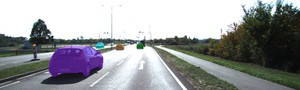}\\
  \includegraphics[width=0.23\textwidth,frame]{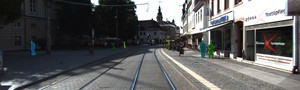}\hspace{1px}%
  \includegraphics[width=0.23\textwidth,frame]{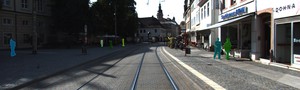}\hspace{1px}%
  \includegraphics[width=0.23\textwidth,frame]{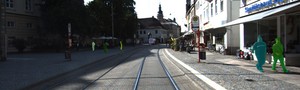}\hspace{1px}%
  \includegraphics[width=0.23\textwidth,frame]{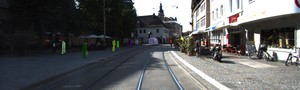}\\%
\includegraphics[width=0.23\textwidth,frame]{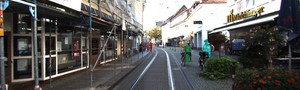}\hspace{1px}%
  \includegraphics[width=0.23\textwidth,frame]{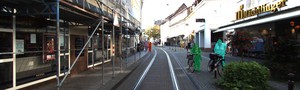}\hspace{1px}%
  \includegraphics[width=0.23\textwidth,frame]{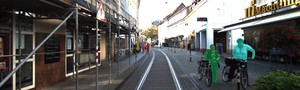}\hspace{1px}%
  \includegraphics[width=0.23\textwidth,frame]{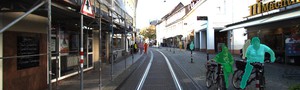}\\%
  \includegraphics[width=0.23\textwidth,frame]{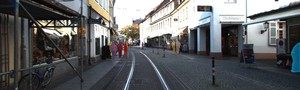}\hspace{1px}%
  \includegraphics[width=0.23\textwidth,frame]{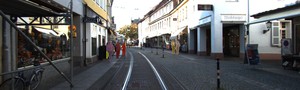}\hspace{1px}%
  \includegraphics[width=0.23\textwidth,frame]{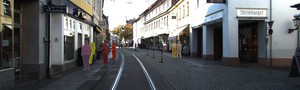}\hspace{1px}%
  \includegraphics[width=0.23\textwidth,frame]{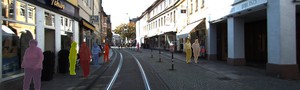}\\%
  \includegraphics[width=0.23\textwidth,frame]{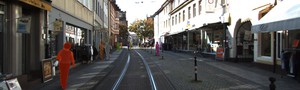}\hspace{1px}%
  \includegraphics[width=0.23\textwidth,frame]{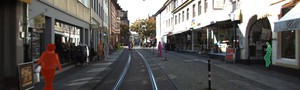}\hspace{1px}%
  \includegraphics[width=0.23\textwidth,frame]{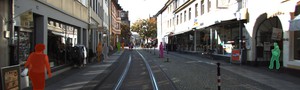}\hspace{1px}%
  \includegraphics[width=0.23\textwidth,frame]{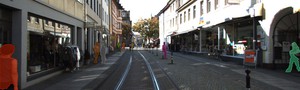}\\%
  \includegraphics[width=0.23\textwidth,frame]{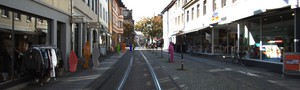}\hspace{1px}%
  \includegraphics[width=0.23\textwidth,frame]{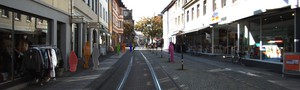}\hspace{1px}%
 \includegraphics[width=0.23\textwidth,frame]{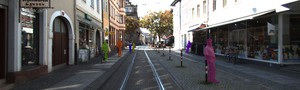}\hspace{1px}%
 \includegraphics[width=0.23\textwidth,frame]{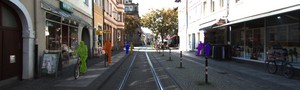}\\

\raggedleft
\begin{tikzpicture}[node distance=2cm]
\node (A) at (2.75, 0) {};
\node (B) at (13.0, 0) {};
\draw[-{Stealth}, to path={-- (\tikztotarget)}](A) edge (B);
\node[text width=1cm] at (13.5,0){time};
\end{tikzpicture}
\vspace{-2px}
  \caption{\textbf{Additional qualitative results on KITTI-MOTS.} Our method successfully tracks and segments cars and pedestrians in automotive scenarios, even when observed from a large distance (\textit{sixth row from the bottom}) and bridges occlusions (\textit{fifth row}).}
  
  \label{fig:qualitative_mots}
\end{figure}

In this section, we provide additional qualitative results on the validation split of all three datasets, DAVIS'19 \cite{Caelles19arXiv} in Fig. \ref{fig:qualitative_davis}, YouTube-VIS \cite{Yang19ICCV} in Fig. \ref{fig:qualitative_ytvis} and KITTI-MOTS \cite{Voigtlaender19CVPR} in Fig. \ref{fig:qualitative_mots}. As can be seen, our method can reliably segment and track a large variety of objects in diverse scenarios, and is fairly robust to scale changes and brief occlusions.

\clearpage

\end{document}